\documentclass{article} % For LaTeX2e
\usepackage{iclr2026_conference,times}

\usepackage{amsmath,amsfonts,bm}

\def\eqref#1{equation~\ref{#1}}
\def\1{\bm{1}}

\DeclareMathAlphabet{\mathsfit}{\encodingdefault}{\sfdefault}{m}{sl}
\SetMathAlphabet{\mathsfit}{bold}{\encodingdefault}{\sfdefault}{bx}{n}

\usepackage{hyperref}
\usepackage{url}

\usepackage{wrapfig}
\usepackage{algorithm}
\usepackage{algpseudocode}

\usepackage{graphicx}
\usepackage{amsmath} 
\usepackage{xcolor} % Required for defining custom colors
\usepackage{tcolorbox} % Loads the main tcolorbox package
\usepackage{booktabs}
\usepackage[table]{xcolor}
\usepackage{multirow}
\usepackage{hyperref}
\usepackage{cleveref}
\usepackage{amssymb}
\usepackage{float}
\usepackage{caption}

\usepackage{listings}
\tcbuselibrary{listings}
\lstdefinestyle{prompt}{
  basicstyle=\ttfamily\footnotesize,
  breaklines=true,
  columns=fullflexible,
  keepspaces=true,
  showstringspaces=false
}

\newtcblisting{promptbox}[2][]{
  title={#2},
  colback=gray!3,
  colframe=black!60,
  boxrule=0.5pt,
  arc=1mm,
  listing only,
  listing options={style=prompt},
  #1
}

\newtcolorbox{comparisonbox}{
  colback=gray!3,
  colframe=black!55,
  boxrule=0.6pt,
  arc=2pt,
  left=5pt,
  right=5pt,
  top=5pt,
  bottom=5pt
}

\newif\ifcomments
\commentsfalse    % Hide comments
\ifcomments
  \newcommand{\kushan}[1]{\textcolor{blue}{[Kushan: #1]}}
  \newcommand{\ej}[1]{\textcolor{orange}{[EJ: #1]}}
  \newcommand{\hannah}[1]{\textcolor{magenta}{[Hannah: #1]}}
  \newcommand{\estevam}[1]{\textcolor{purple}{[Estevam: #1]}}
\else
  \newcommand{\kushan}[1]{}
  \newcommand{\ej}[1]{}
  \newcommand{\hannah}[1]{}
  \newcommand{\estevam}[1]{}
\fi

\newcommand{\ours}[1]{\textsc{HypReflect}{}}
\newcommand{\ourstoo}[1]{\textsc{HypReflect+Sum}{}}
\newcommand{\reflectionhelp}[1]{\textcolor{blue!70!black}{\textbf{#1}}}
\newcommand{\refinementissue}[1]{\textcolor{red!70!black}{\textbf{#1}}}

\title{Hypotheses-Guided Self Distillation \\for Continual Personalization}

\author{
\textbf{EunJeong Hwang}\textsuperscript{1}\thanks{Work done as an intern at Megagon Labs.} ,
\textbf{Kushan Mitra}\textsuperscript{2},
\textbf{Dan Zhang}\textsuperscript{2},
\textbf{Hannah Kim}\textsuperscript{2},
\textbf{Estevam Hruschka}\textsuperscript{2} \\
\textsuperscript{1}University of British Columbia \quad \textsuperscript{2}Megagon Labs, USA \\
\texttt{ejhwang@cs.ubc.ca} \quad \texttt{\{kushan,dan\_z,hannah,estevam\}@megagon.ai}
}

\iclrfinalcopy % Uncomment for camera-ready version, but NOT for submission.
\begin{document}

\maketitle
\begin{abstract}
As people increasingly interact with LLM assistants in daily life, continually adapting to individual preferences has become essential for effective long-term interactions. However, user preferences are rarely stated in full, and instead emerge through heterogeneous, latent, and noisy signals, with existing methods relying on raw interaction histories or costly reward-based optimization to manage personalization. We introduce \ours{}, a reliable, scalable framework for continual personalization that infers explicit, uncertainty-aware \kushan{We don't use uncertainty-aware anywhere else. Can we make this 'weighted' preference hypotheses?} preference hypotheses from diverse user signals, reflectively refines them as new evidence accumulates, and incorporates the resulting user model through hypotheses-guided self-distillation.
Experiments across three personalization settings: online personalization, multi-session interactions, and implicit behavioral signals, show that \ours{} outperforms a range of baselines, including raw-history and incremental-update methods. We further demonstrate strong generalization to unseen users and cross-domain settings, along with stability across context budgets, reusable hypotheses, and more focused personalization. These results suggest a step towards reliable and scalable continual personalization through explicit, revisable user preference hypotheses.\footnote{Code will be released upon acceptance.}

\end{abstract}

\begin{wrapfigure}[26]{R}{0.5\linewidth}
    \centering
    \includegraphics[width=\linewidth, trim=1cm 1cm 7cm 1cm, clip]{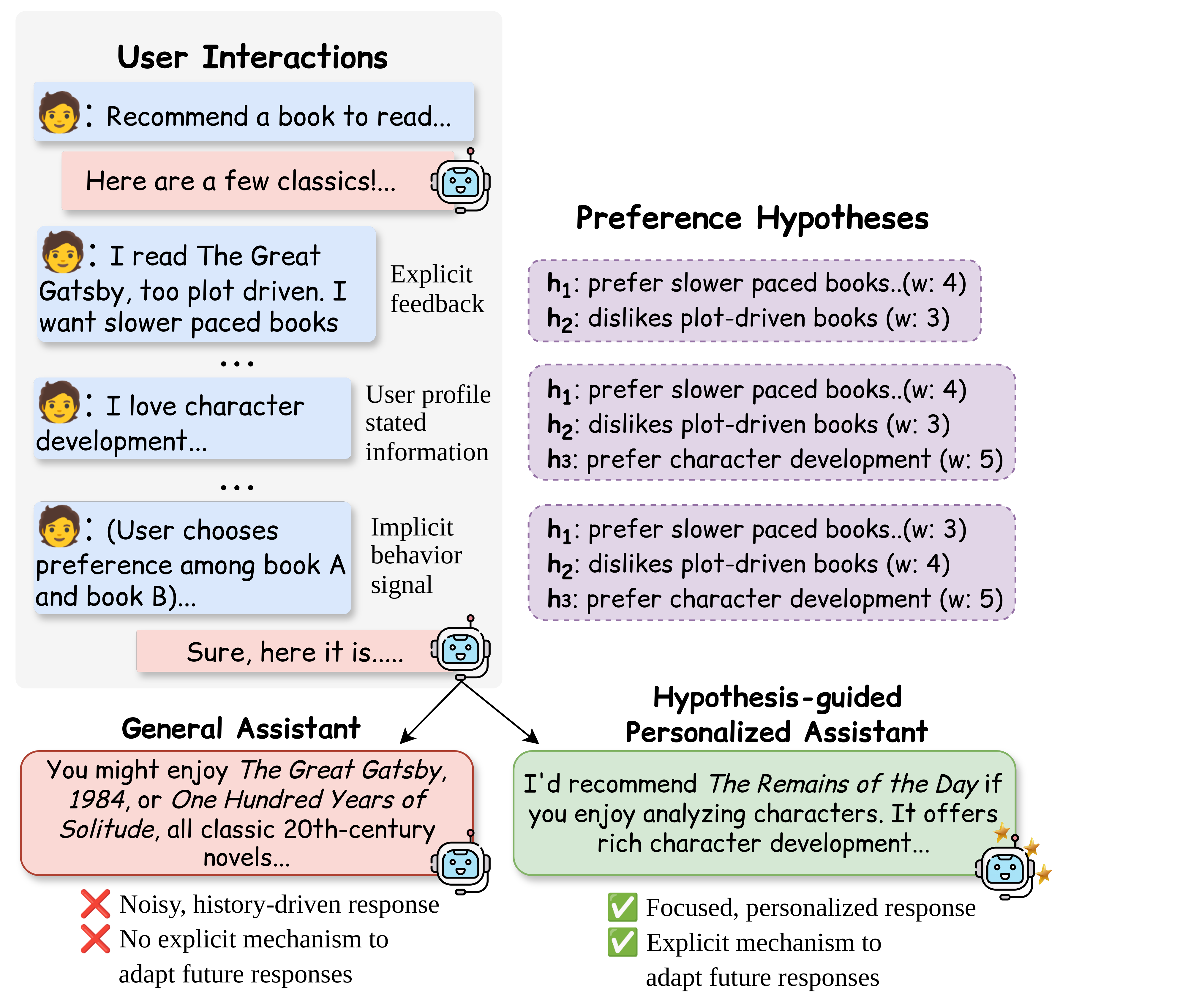}
    \caption{Preference hypotheses are generated across personalization settings from explicit feedback, implicit behavioral signals, and user profile information, along with confidence weights. They maintain explicit, revisable beliefs about the user and guide models to produce personalized responses.}
    \label{fig:motivation}
    \vspace{-10pt}
\end{wrapfigure}
\section{Introduction}
Conversational assistants that continuously adapt to individual users can provide increasingly effective and personalized support in diverse settings, including writing assistance~\citep{mysore-etal-2024-pearl}, conversational recommendation~\citep{liang-etal-2024-llm, kim-etal-2025-towards, zhu-etal-2025-llm-based}, and education~\citep{10.1145/3701716.3717527}. 
Personalization allows assistants to better address a user's needs~\citep{zhang-etal-2018-personalizing, salemi-etal-2024-lamp, zhang2025personalizationlargelanguagemodels, zhao-etal-2025-personalens}, while continual adaptation enables long-term interactions with users by improving relevance, as assistants develop a richer understanding of users over time~\citep{xu-etal-2022-beyond, 10.1609/aaai.v38i17.29946, li-etal-2025-hello}.

However, users rarely articulate their preferences in full and may not always be consciously aware of them, making continual adaptation inherently challenging.
Figure~\ref{fig:motivation} illustrates how evidence about a user's preferences emerges incrementally from explicit feedback, implicit behavior, and user-profile information, requiring it to be consolidated into an explicit representation that can be revised as new evidence arrives. These signals are often sparse, noisy, and context dependent, making it difficult to distinguish stable preferences from task-specific requirements. Moreover, evidence may be distributed across long multi-turn and multi-session histories~\citep{maharana-etal-2024-evaluating, wu2025longmemevalbenchmarkingchatassistants}, which language models often struggle to track consistently as conversations grow~\citep{laban2025llmslostmultiturnconversation}. The central problem is therefore to transform an evolving stream of interactions into a user representation that is both stable and revisable.

Existing approaches personalize LLMs by conditioning on user profiles, retrieved memories, or raw interaction histories~\citep{packer2024memgptllmsoperatingsystems, zhao2025llmsrecognizepreferencesevaluating, li2026horizonbenchlonghorizonpersonalizationevolving}, making it increasingly costly to retain and retrieve past interactions over time. Other methods infer user traits or summaries through RL but require reward models, making continual optimization costly~\citep{wan2025enhancingpersonalizedmultiturndialogue, nam2026learningsummarizeuserinformation}. \citet{buening2026aligning} applies self-distillation, improving scalability for continual personalization, but relies on raw interaction histories, leaving it unclear which preferences are learned and whether they generalize to future contexts.

In this work, we introduce \ours{}, a scalable framework for continual personalization that maintains \emph{preference hypotheses} i.e. explicit, uncertainty-aware, and revisable beliefs about the user. \ours{} distills diverse user signals (e.g., Fig.~\ref{fig:motivation}) into preference hypotheses, reflectively consolidates useful user signals across interactions, and incorporates the refined hypotheses to guide personalized response generation through hypotheses-guided self-distillation.

We evaluate \ours{} across three personalization settings covering explicit feedback, multi-session interactions, and implicit behavioral choices, against a comprehensive suite of baselines, including self-distillation over raw interaction histories, incrementally updated summaries and hypotheses, and reflective prompting. Across these settings, \ours{} achieves relative improvements of up to 4.4\%, 10.3\%, and 5.0\% respectively\footnote{Maximum relative improvement per dataset with \ourstoo{} in Table~\ref{tab:main-results}.}, while generalizing effectively to unseen users, shifted user characteristics, and cross-domain personalization. Further analyses demonstrate robustness across context budgets, hypothesis reusability, and more focused personalization. Overall, our findings show that reflective refinement enables stable and revisable preference hypotheses over long interaction histories, and that models benefit substantially from explicit, reusable user representations, enabling reliable and scalable continual personalization.
% Overall, our findings show that models benefit substantially from maintaining explicit, reusable user representations, enabling stable and scalable continual personalization.

% \input{sections/introduction}
\section{Problem Formulation}
\label{sec:hypothesis_formulation}
\paragraph{Continual Personalization} considers an assistant that repeatedly interacts with a user. At interaction $t$, the assistant observes the preceding conversation history $C_{<t}$ and the current user turn $u_t$, which may contain feedback $f_{t-1}$ on the preceding assistant response, a new request $q_t$, or both. The assistant then generates a response:
% \vspace{-0.15cm}
\[y_t \sim \pi_{\theta}(\cdot \mid C_{<t}, u_t).\]
As interactions accumulate, the history provides increasing signals about the user's preferences. The goal therefore is to continually infer, maintain, and use this understanding to personalize future responses.

\paragraph{Self-Distillation for Continual Personalization.}
Self-distillation uses the same model as both a student and a teacher, with the teacher conditioned on richer information. For continual personalization, self-distillation offers a simple and scalable alternative to costly reward optimization by treating the subsequent user turn $u_{t+1} = \{f_t, q_{t+1}\}$ as privileged information when ground-truth personalized responses are unavailable~\citep{buening2026aligning}.
The student and teacher distributions are
\begin{equation*}
\pi_t^{\mathrm{S}}=
\pi_\theta(\cdot \mid C_{<t}, u_t), \text{ }
\pi_t^{\mathrm{T}}=
\pi_\theta(\cdot \mid C_{<t}, u_t, u_{t+1}),
\end{equation*}
and the teacher's feedback-informed predictions are distilled into the student by minimizing
\begin{equation*}
\mathcal{L}_{\mathrm{SD}}(\theta)
=
\sum_t
D_{\mathrm{KL}}\left(
\pi_t^{\mathrm{S}}\middle\Vert
\operatorname{stopgrad}\left[\pi_t^{\mathrm{T}}\right]
\right).
\label{eq:self-distillation}
\end{equation*}
This enables learning from preferences revealed by subsequent interactions without an external teacher or reward model. However, directly internalizing preferences from raw interaction histories may not generalize well beyond the current interaction.

\paragraph{Preference Hypotheses Representation.}
Hence, we introduce an explicit abstraction layer between raw interactions and response generation. We represent the model's understanding of the user as a set of preference hypotheses $H_t$ and formulate personalized response generation as
\[
y_t \sim \pi_{\theta}(\cdot \mid C_{<t}, q_t, H_t).
\]
This representation makes inferred preferences explicit, reusable, and revisable as new evidence arrives~\citep{liu2026textuniversalinterfacetransferable}. The key challenge here is how to construct hypotheses from local interaction signals, refine and maintain them over long histories, and use them to guide self-distillation.

\section{\ours{}: Continual User Modeling through Reflective Refinement}
\label{sec:methodology}

\begin{figure*}
    \centering
    \includegraphics[width=1.0\linewidth, trim=0cm 0.2cm 0cm 0cm, clip]{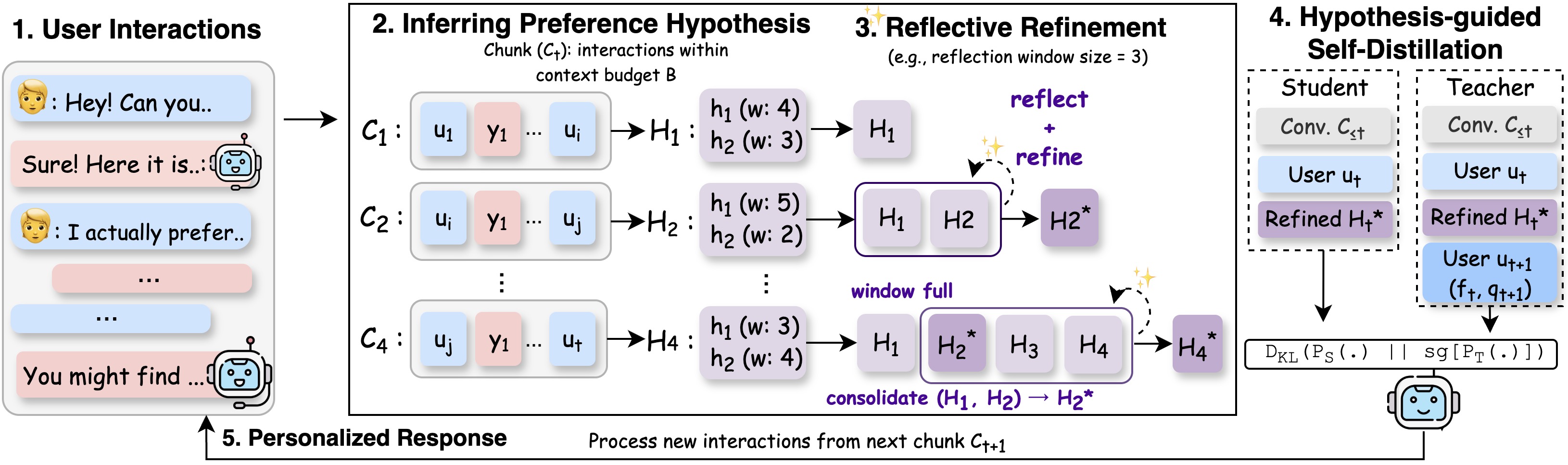}
    \caption{Overview of \ours{}. Preference hypotheses are generated from context-bounded chunks (Steps 1–2) and reflectively refined over time (Step 3). The refined user state conditions both student and teacher during self-distillation, with the teacher seeing the next user turn as privileged information (Step 4). The resulting model produces continually personalized responses (Step 5).}
    \label{fig:methodology-figure}
\end{figure*}

To support continual personalization, we propose \ours{}, a framework that introduces an explicit preference hypothesis layer to better understand and respond to the user.
Our framework consists of three stages. 
First, the model infers preference hypotheses from individual interaction chunks, capturing localized signals about the user (\S{}\ref{sec:hyp-inference}). Second, it recurrently refines these hypotheses by reflecting and consolidating user signals across chunks (\S{}\ref{sec:reflection}).
Finally, the refined hypotheses guide self-distillation, enabling the model to use subsequent user signals as privileged supervision for learning personalized response behavior (\S{}\ref{sec:self-distillation}). We also consider a summary-augmented variant that jointly maintains summaries and hypotheses.

\subsection{Inferring Preference Hypotheses from Interactions}
\label{sec:hyp-inference}

\paragraph{Hypotheses representation.}
To capture diverse interaction signals from the user (e.g., Fig.~\ref{fig:motivation}), we represent the model's understanding of the user as a revisable set of preference hypotheses $H_t$, comprising $m$ multiple candidate hypotheses paired with verbalized confidence scores: 

\vspace{-0.4cm}
\[
H_t = \{h_{t1}, \ldots, h_{tm}\},
\qquad
h_{tj} = (s_{tj}, w_{tj}),
\]
where $s_{tj}$ is a natural-language statement describing a potential user preference and $w_{tj} (\in \{1,\cdots,5\})$ denotes the model's confidence in that statement. Maintaining multiple hypotheses preserves alternative beliefs that can be revised as new evidence arrives, while confidence distinguishes well-supported preferences from uncertain ones. The natural-language representation makes the hypotheses interpretable, revisable, and directly usable for personalized response generation.

\paragraph{Inferring local hypotheses from interaction chunks.}
Inferring preferences from the full interaction history is often impractical due to context-length limitations and, in online settings, because future interactions are not yet available. We therefore partition the observed interaction history into consecutive chunks $\{C_1,\ldots,C_N\}$, each containing the longest sequence of complete interactions within a token budget $B$. For each chunk $C_i$, we infer a set of \emph{local hypotheses} (Step 2 in Fig.~\ref{fig:methodology-figure})  as
% \vspace{-0.2cm}
\begin{equation*}
H_t=\textsc{GenerateHypotheses}(C_t)=\{h_{tj}\}_{j=1}^{m}.
\end{equation*}

The model is instructed to infer reusable preferences grounded in the user's instructions, corrections, feedback, and choices, while avoiding interaction-specific restatements and unsupported generalizations.

We additionally define a summary-augmented variant, where we generate a local summary $S_i$ for each interaction chunk. 
Summaries preserve relevant conversational context from $C_i$ that grounds the hypotheses, whereas $H_i$ captures reusable preferences that are explicitly stated or inferred from user signals.

\subsection{Maintaining User Understanding through Reflective Refinement}
\label{sec:reflection}

As interactions accumulate, the model must maintain preference hypotheses that (1) \textit{explain past interaction histories} and (2) \textit{predict future user behavior}. Simply using the most recent hypotheses discards earlier preferences, whereas repeatedly reconsidering the full interaction history is computationally infeasible. A refinement mechanism should therefore preserve long-term preferences while remaining revisable and bounded in computational context size.

A simple approach is incremental updating, which iteratively refines a single running user state. However, since each update operates on an already compressed representation, over time it is either prone to error accumulation, or it may have valid earlier preferences overwritten by recent observations. We empirically show that this leads to unstable long-term personalization (\S{}\ref{subsec:main-results}).

\paragraph{Reflective refinement across chunks.}
To preserve evidence across multiple chunks, without relying exclusively on a recursively compressed state, we propose \textbf{reflective refinement} (Step 3 in Fig.~\ref{fig:methodology-figure}). Rather than immediately merging each new local hypothesis set into a single running state, the model jointly reconsiders multiple independently generated local sets. By comparing hypotheses across interaction chunks, the model consolidates consistent evidence, resolves conflicts, merges redundant hypotheses, and revises beliefs that are no longer supported.

To keep refinement tractable, we maintain a bounded reflection window. Let $H_i$ denote the local hypothesis set inferred from chunk $C_i$, $H_i^{*}$ the refined hypothesis set after observing the first $i$ chunks, and $W$ the reflection-window size. We construct the refinement input as

\begin{equation*}
\mathcal{I}_i=
\begin{cases}
[H_1,\ldots,H_i], & i\leq W, \\[4pt]
[H_{i-W+1}^{*},H_{i-W+2},\ldots,H_i], & i>W,
\end{cases}
\label{eq:reflection_input}
\vspace{+5pt}
\end{equation*}
and update the maintained hypothesis set as
$H_i^{*}=\textsc{ReflectAndRefine}(\mathcal{I}_i)$.

When $i \leq W$, the model reflects over all available local hypothesis sets. Thereafter, it replaces older local sets with their refined representation while retaining the most recent $W-1$ local sets. This keeps the refinement input bounded without discarding earlier preference information. 

For our summary-augmented variant, we apply the same windowed recurrent refinement procedure to a jointly maintained summary and hypothesis set, $Z_i=(S_i,H_i)$, for each chunk, producing $Z_i^{*}=(S_i^{*},H_i^{*})$. The refined summary $S_i^{*}$ consolidates the context used to ground and revise the hypotheses, while $H_i^{*}$ maintains the refined preferences used for future personalization.

\subsection{Hypothesis-Guided Self-Distillation}
\label{sec:self-distillation}
To effectively use the refined user state for personalized response generation, we incorporate it into self-distillation, training the model to translate maintained preferences into personalized responses. Building on the formulation in~\S{}\ref{sec:hypothesis_formulation}, we condition both the teacher and student on the same refined state, providing a shared understanding of the user and making the teacher's targets learnable by the student. The teacher additionally observes the subsequent user turn $u_{t+1}=(f_t,q_{t+1})$ and uses it as privileged hindsight to generate a preference-grounded response.

At interaction $t$, given the preceding history $C_{<t}$, current user turn $u_t$, refined user hypotheses $H_t^{*}$, and subsequent user turn $u_{t+1}$, the student and teacher distributions are
\begin{equation*}
\begin{aligned}
P_S(y_t)
&=
\pi_{\theta}\left(
y_t \mid C_{<t},u_t,H_t^{*}
\right), \text{ }
P_T(y_t)
=
\pi_{\theta}\left(
y_t \mid C_{<t},u_t,H_t^{*},u_{t+1}
\right).
\end{aligned}
\end{equation*}

The teacher uses $u_{t+1}$ to provide feedback-informed supervision. We then minimize the reverse KL divergence from the student to the detached teacher:
% \vspace{-0.2cm}
\begin{equation*}
\mathcal{L}_{\mathrm{distill}}
=
\mathbb{E}_{t}
\left[
D_{\mathrm{KL}}
\left(
P_S(\cdot)
\,\Vert\,
\operatorname{stopgrad}[P_T(\cdot)]
\right)
\right].
\end{equation*} The objective is computed token-wise over the generated response (Step 4 in Fig.~\ref{fig:methodology-figure}).

\section{Experimental Setup}

\subsection{Datasets and Evaluation}
To test whether models can infer reusable user preferences from diverse interaction signals and use them to personalize future responses, we evaluate three complementary settings: continual adaptation from explicit user feedback, multi-session personalization using conversational user profiles, and preference inference from implicit behavioral signals. See Appendix~\ref{appendix:dataset} and \ref{app:eval-detail} more details.

\paragraph{HelpSteer2}\citep{wang2024helpsteer}.
To evaluate continual adaptation from \textbf{explicit user feedback}, we construct an online personalization setting from HelpSteer2. Each user is associated with a set of writing-style preferences, and across multiple interactions, a user simulator provides feedback on one or two relevant preference dimensions after each model response. \\\texttt{Evaluation} Models are trained over 250 interactions and evaluated every 50 interactions on held-out prompts using an LLM judge. We report pairwise win rates against the base model with respect to the target user's preferences.

\paragraph{HiCupid}\citep{mok-etal-2025-exploring}.
To test personalization from \textbf{user profile information}, we use HiCupid, where user preferences and attributes are stated naturally across multi-session conversations. We train a model on QA interactions from 300 users and evaluate whether it can use previous user information to personalize future responses. \\\texttt{Evaluation} We evaluate both seen-user and unseen-user generalization using held-out prompts and report pairwise win rates from an LLM judge\footnote{We use \texttt{deepseek-v4-flash} as the LLM judge and verify its consistency with other judges in Appendix~\ref{app:llm-judge-validity}.} assessing personalization and response quality.

\paragraph{Flight Recommendation}\citep{qiu2026bayesian}.
To evaluate preference inference from \textbf{implicit behavioral signals}, we use a Flight Recommendation task. We utilize multi-turn recommendation trajectories from 624 users, with up to 25 interactions per user, where user choices provide partial evidence about latent preferences. These preferences may include non-obvious trade-offs, requiring models to infer user-specific behavior rather than rely on fixed assumptions. 
\\\texttt{Evaluation} We report final-round preference prediction accuracy on the original flight task, held-out feature settings, and cross-domain hotel transfer.

\subsection{Model Variants and Training Setup}
(1) \textbf{\textit{Base.}} Uses the vanilla language model without fine-tuning, conditioned on the relevant raw interaction history, to generate a response.
(2) \textbf{\textit{SD.}}
Self-distillation is performed directly on the raw interaction history. When the history exceeds the context budget, only the most recent portion is retained;
(3) \textbf{\textit{SD-IncSum.}}
The model is trained with an incrementally updated summary: a running summary $S_i$ is maintained and refreshed after each chunk;
(4) \textbf{\textit{SD-IncHyp.}}
Instead of a summary, the model maintains a running hypothesis set $H_i$, updated with the same incremental strategy;
(5) \textbf{\textit{\ours{}.}} 
Local hypotheses $H_i$ are first generated from individual chunks and then consolidated into a refined set $H_i^{*}$ through the reflective refinement process described in \S{}\ref{sec:reflection};
(6) \textbf{\textit{\ourstoo{}.}}
Reflective refinement is applied to the paired memory state $Z_i=(S_i,H_i)$, yielding $Z_i^{*}=(S_i^{*},H_i^{*})$, to examine whether summaries supply complementary information during hypothesis refinement.
\paragraph{Training setup.} 
% \kushan{Depending on space, this set up can be described in appendix as well.}\ej{Some lora details can go into appendix, but we need to describe what models we used in main text.}
We use Qwen3.5-4B, Qwen3.5-9B~\citep{yang2025qwen3}, and Gemma4-4B~\citep{team2026gemma} as backbone LLMs. Additional hyperparameter details are provided in Appendix~\ref{app:train-details}.
% \kushan{Missing reference, pointing to table for now}
% Across all experiments, we use LoRA with rank 32 and scaling factor $\alpha=64$, and train using bfloat16 precision.

\section{Results}
\begin{table*}[h]
\centering
\caption{Performance on three continual-personalization benchmarks covering continual adaptation with explicit user feedback (HelpSteer2), user-profile information across multiple sessions (HiCupid), and implicit behavioral signals (Flight). We report win rate against Base for HelpSteer2 (averaged over 50--250 interactions) and HiCupid, and accuracy for Flight.}
\resizebox{\textwidth}{!}{%
\begin{tabular}{l|ccc|ccc|ccc}
\toprule
& \multicolumn{3}{c|}{HelpSteer2 (Win Rate $\uparrow$)}
& \multicolumn{3}{c|}{HiCupid (Win Rate $\uparrow$)}
& \multicolumn{3}{c}{Flight Recommendation (Acc. $\uparrow$)} \\
\cmidrule(lr){2-4}
\cmidrule(lr){5-7}
\cmidrule(lr){8-10}
% &\multicolumn{3}{c|}{Win Rate $\uparrow$}
% & \multicolumn{3}{c|}{Win Rate $\uparrow$}
% & \multicolumn{3}{c}{Accuracy $\uparrow$} \\
Method
& Gemma4-4B & Qwen3.5-4B & Qwen3.5-9B
& Gemma4-4B & Qwen3.5-4B & Qwen3.5-9B
& Gemma4-4B & Qwen3.5-4B & Qwen3.5-9B \\
\midrule
Base
& -- & -- & --
& -- & -- & --
& 40.9 & 40.3 & 41.7 \\
\midrule
SD
& 87.0 & 84.5 & 89.0
& 57.3 & 69.7 & 63.3
& 61.8 & 73.8 & 74.4 \\
SD-IncSum
& 85.8 \textcolor{red}{\scriptsize $(-1.2\downarrow)$}
& 81.0 \textcolor{red}{\scriptsize $(-3.5\downarrow)$}
& 86.6 \textcolor{red}{\scriptsize $(-2.4\downarrow)$}
& 58.4 \textcolor{teal}{\scriptsize $(+1.1\uparrow)$}
& 67.9 \textcolor{red}{\scriptsize $(-1.8\downarrow)$}
& \underline{67.7} \textcolor{teal}{\scriptsize $(+4.4\uparrow)$}
& 50.5 \textcolor{red}{\scriptsize $(-11.3\downarrow)$}
& \underline{74.4} \textcolor{teal}{\scriptsize $(+0.6\uparrow)$}
& 67.0 \textcolor{red}{\scriptsize $(-7.4\downarrow)$} \\
SD-IncHyp
& \underline{89.6} \textcolor{teal}{\scriptsize $(+2.6\uparrow)$}
& 81.9 \textcolor{red}{\scriptsize $(-2.6\downarrow)$}
& 88.2 \textcolor{red}{\scriptsize $(-0.8\downarrow)$}
& 60.7 \textcolor{teal}{\scriptsize $(+3.4\uparrow)$}
& \underline{71.6} \textcolor{teal}{\scriptsize $(+1.9\uparrow)$}
& 66.2 \textcolor{teal}{\scriptsize $(+2.9\uparrow)$}
& 57.3 \textcolor{red}{\scriptsize $(-4.5\downarrow)$}
& \textbf{74.8} \textcolor{teal}{\scriptsize $(+1.0\uparrow)$}
& \underline{75.4} \textcolor{teal}{\scriptsize $(+1.0\uparrow)$} \\
\midrule
\rowcolor{gray!15}
\ours{}
& 86.3 \textcolor{red}{\scriptsize $(-0.7\downarrow)$}
& \textbf{88.9} \textcolor{teal}{\scriptsize $(+4.4\uparrow)$}
& \underline{90.3} \textcolor{teal}{\scriptsize $(+1.3\uparrow)$}
& \underline{61.1} \textcolor{teal}{\scriptsize $(+3.8\uparrow)$}
& 71.1 \textcolor{teal}{\scriptsize $(+1.4\uparrow)$}
& \underline{67.7} \textcolor{teal}{\scriptsize $(+4.4\uparrow)$}
& \underline{62.3} \textcolor{teal}{\scriptsize $(+0.5\uparrow)$}
& 74.0 \textcolor{teal}{\scriptsize $(+0.2\uparrow)$}
& \textbf{76.0} \textcolor{teal}{\scriptsize $(+1.6\uparrow)$} \\
\rowcolor{gray!15}
\ours{}+Sum
& \textbf{90.0} \textcolor{teal}{\scriptsize $(+3.0\uparrow)$}
& \underline{88.2} \textcolor{teal}{\scriptsize $(+3.7\uparrow)$}
& \textbf{92.1} \textcolor{teal}{\scriptsize $(+3.1\uparrow)$}
& \textbf{61.9} \textcolor{teal}{\scriptsize $(+4.6\uparrow)$}
& \textbf{71.8} \textcolor{teal}{\scriptsize $(+2.1\uparrow)$}
& \textbf{69.8} \textcolor{teal}{\scriptsize $(+6.5\uparrow)$}
& \textbf{64.9} \textcolor{teal}{\scriptsize $(+3.1\uparrow)$}
& 74.2 \textcolor{teal}{\scriptsize $(+0.4\uparrow)$}
& 75.3 \textcolor{teal}{\scriptsize $(+0.9\uparrow)$} \\
\bottomrule
\end{tabular}
}
\label{tab:main-results}
\vspace{-10pt}
\end{table*}

\subsection{Main Results}
\label{subsec:main-results}
\paragraph{Preference hypotheses provide a stronger representation than summaries.} 
Table~\ref{tab:main-results} compares performance across three personalization benchmarks. Under the incremental update strategy, \textsc{SD-IncHyp} outperforms \textsc{SD-IncSum} in eight of nine settings, showing that abstracting interaction signals into reusable preference beliefs is generally more effective than retaining them as summaries.
\ours{} further improves over \textsc{SD-IncHyp} and outperforms \textsc{SD} in eight of nine settings, indicating that the effectiveness of preference hypotheses also depends on maintaining them accurately and consistently over time.

\paragraph{Reflective refinement makes personalization more reliable.} Furthermore, \ours{} outperforms \textsc{SD} in eight of nine settings and \ourstoo{} improves over it in all nine. \ourstoo{} also outperforms \ours{} in seven settings, suggesting that summaries provide complementary evidence omitted during hypothesis abstraction and ground refinement in both observed interactions and inferred preferences.

In contrast, \textsc{SD-IncSum} and \textsc{SD-IncHyp} underperform \textsc{SD} in six and three settings, respectively.
We find that incremental updates can propagate earlier errors and allow recent observations to overwrite valid prior information, resulting in unstable supervision over time. In contrast, reflective refinement maintains more reliable preference hypotheses as interaction histories grow.

\begin{wraptable}[11]{R}{0.6\linewidth}
\centering
\scriptsize
\caption{Comparison of \ours{} against the base model and Base+RefHyp (base model reflective prompting approach).}
\resizebox{0.95\linewidth}{!}{%
\begin{tabular}{lcc|ccc}
\toprule
& \multicolumn{2}{c|}{\textbf{HiCupid}}
& \multicolumn{3}{c}{\textbf{Flight Recommendation}} \\
\cmidrule(lr){2-3}
\cmidrule(lr){4-6}
& \multicolumn{2}{c|}{Win Rate of \ours{} $\uparrow$}
& \multicolumn{3}{c}{Accuracy $\uparrow$} \\
Model
& vs.\ Base
& vs.\ Base+RefHyp
& Base
& Base+RefHyp
& \ours{} \\
\midrule
Gemma4-4B
& 61.1
& 59.7
& 40.9
& 47.9
& \textbf{62.3} \\

Qwen3.5-4B
& 71.1
& 62.2
& 40.3
& 42.0
& \textbf{74.0} \\

Qwen3.5-9B
& 67.7
& 57.6
& 41.7
& 46.8
& \textbf{76.0} \\
\bottomrule
\end{tabular}%
}
\label{tab:sd-vs-prompting}
\end{wraptable}

\paragraph{Reflective hypotheses prompting yields consistent gains, which self-distillation further amplifies.}
Table~\ref{tab:sd-vs-prompting} demonstrates the effectiveness of reflective refinement both as an inference-time prompting method and when combined with self-distillation. We implement \textsc{Base+RefHyp} similarly to \ours{}, reflectively refining preference hypotheses to generate a response but without self-distillation.
Even when used only through prompting, reflectively refining preference hypotheses improves the base model, with \textsc{Base+RefHyp} gaining an average of 4.6 accuracy points on Flight Recommendation and narrowing the gap to \ours{} on HiCupid, reducing the average win rate of \ours{} from 66.6\% against \textsc{Base} to 59.8\%. Combining reflective refinement with hypothesis-guided self-distillation yields substantially larger gains, as \ours{} outperforms \textsc{Base+RefHyp} by an average of 25.2 accuracy points on Flight Recommendation and achieves an average win rate of 59.8\% against it on HiCupid. These results show that reflective refinement is effective even as inference-time prompting and combining it with self-distillation more effectively incorporates user preferences into personalized response behavior.

\begin{table*}[t]
\centering
\small
\caption{Generalization results. Flight$\rightarrow$Features: robustness across 2--8 features (with models trained on 4 features). Flight$\rightarrow$Hotel: cross-domain transfer. HiCupid Unseen: performance on unseen users.}
\resizebox{\textwidth}{!}{%
\begin{tabular}{l|ccc|ccc|ccc}
\toprule
& \multicolumn{3}{c|}{Flight $\rightarrow$ Features (Acc. $\uparrow$)}
& \multicolumn{3}{c|}{Flight $\rightarrow$ Hotel (Acc. $\uparrow$)}
& \multicolumn{3}{c}{HiCupid Unseen (Win Rate $\uparrow$)} \\
\cmidrule(lr){2-4}
\cmidrule(lr){5-7}
\cmidrule(lr){8-10}
% & \multicolumn{3}{c|}{Accuracy $\uparrow$}
% & \multicolumn{3}{c|}{Accuracy $\uparrow$}
% & \multicolumn{3}{c}{Win Rate $\uparrow$} \\
Method
& Gemma4-4B & Qwen3.5-4B & Qwen3.5-9B
& Gemma4-4B & Qwen3.5-4B & Qwen3.5-9B
& Gemma4-4B & Qwen3.5-4B & Qwen3.5-9B \\
\midrule
SD
& 55.4 & 64.9 & 63.7
& 54.8 & 61.8 & 62.0
& 57.0 & 71.7 & 66.9 \\
SD-IncSum
& 47.5 \textcolor{red}{\scriptsize\textit{$(-7.9\downarrow)$}}
& 65.6 \textcolor{teal}{\scriptsize\textit{$(+0.7\uparrow)$}}
& 57.1 \textcolor{red}{\scriptsize\textit{$(-6.6\downarrow)$}}
& 48.0 \textcolor{red}{\scriptsize\textit{$(-6.8\downarrow)$}}
& 63.7 \textcolor{teal}{\scriptsize\textit{$(+1.9\uparrow)$}}
& 57.1 \textcolor{red}{\scriptsize\textit{$(-4.9\downarrow)$}}
& 58.7 \textcolor{teal}{\scriptsize\textit{$(+1.7\uparrow)$}}
& 69.9 \textcolor{red}{\scriptsize\textit{$(-1.8\downarrow)$}}
& 69.4 \textcolor{teal}{\scriptsize\textit{$(+2.5\uparrow)$}} \\
SD-IncHyp
& 53.4 \textcolor{red}{\scriptsize\textit{$(-2.0\downarrow)$}}
& 64.8 \textcolor{red}{\scriptsize\textit{$(-0.1\downarrow)$}}
& 65.5 \textcolor{teal}{\scriptsize\textit{$(+1.8\uparrow)$}}
& 52.8 \textcolor{red}{\scriptsize\textit{$(-2.0\downarrow)$}}
& 62.3 \textcolor{teal}{\scriptsize\textit{$(+0.5\uparrow)$}}
& 62.9 \textcolor{teal}{\scriptsize\textit{$(+0.9\uparrow)$}}
& 62.1 \textcolor{teal}{\scriptsize\textit{$(+5.1\uparrow)$}}
& 71.8 \textcolor{teal}{\scriptsize\textit{$(+0.1\uparrow)$}}
& 68.0 \textcolor{teal}{\scriptsize\textit{$(+1.1\uparrow)$}} \\
\midrule
\rowcolor{gray!15}
\ours{}
& 55.6 \textcolor{teal}{\scriptsize\textit{$(+0.2\uparrow)$}}
& \textbf{65.8} \textcolor{teal}{\scriptsize\textit{$(+0.9\uparrow)$}}
& \textbf{66.0} \textcolor{teal}{\scriptsize\textit{$(+2.3\uparrow)$}}
& 55.7 \textcolor{teal}{\scriptsize\textit{$(+0.9\uparrow)$}}
& 63.6 \textcolor{teal}{\scriptsize\textit{$(+1.8\uparrow)$}}
& \textbf{63.8} \textcolor{teal}{\scriptsize\textit{$(+1.8\uparrow)$}}
& \textbf{62.9} \textcolor{teal}{\scriptsize\textit{$(+5.9\uparrow)$}}
& 71.6 \textcolor{red}{\scriptsize\textit{$(-0.1\downarrow)$}}
& 68.1 \textcolor{teal}{\scriptsize\textit{$(+1.2\uparrow)$}} \\
\rowcolor{gray!15}
\ours{}+Sum
& \textbf{57.7} \textcolor{teal}{\scriptsize\textit{$(+2.3\uparrow)$}}
& 65.0 \textcolor{teal}{\scriptsize\textit{$(+0.1\uparrow)$}}
& 65.9 \textcolor{teal}{\scriptsize\textit{$(+2.2\uparrow)$}}
& \textbf{56.5} \textcolor{teal}{\scriptsize\textit{$(+1.7\uparrow)$}}
& \textbf{64.1} \textcolor{teal}{\scriptsize\textit{$(+2.3\uparrow)$}}
& 63.7 \textcolor{teal}{\scriptsize\textit{$(+1.7\uparrow)$}}
& 62.6 \textcolor{teal}{\scriptsize\textit{$(+5.6\uparrow)$}}
& \textbf{73.3} \textcolor{teal}{\scriptsize\textit{$(+1.6\uparrow)$}}
& \textbf{70.3} \textcolor{teal}{\scriptsize\textit{$(+3.4\uparrow)$}} \\
\bottomrule
\end{tabular}
}
\label{tab:generalization}
\vspace{-10pt}
\end{table*}

\subsection{Generalization Results}

\paragraph{Reflective refinement improves robustness under distribution shift.}

Table~\ref{tab:generalization} reports results on three generalization settings. On HiCupid, we evaluate models on users unseen during training. On the Flight Recommendation dataset, we test models trained with 4 features on varying numbers of features (2--8) and a different Hotel domain.

Reflective methods remain the most robust. \ourstoo{} improves over \textsc{SD} in all nine settings, often by substantial margins, and \ours{} outperforms \textsc{SD} in eight out of nine settings. In contrast, \textsc{SD-IncSum} produces large drops on the Flight transfer tasks. \textsc{SD-IncHyp} is more stable and improves performance in most cases, yet still underperforms the reflective approaches on several Flight settings.

These results show that reflective refinement produces higher-quality hypotheses, providing more reliable conditioning signals during self-distillation, allowing the model to learn a cleaner mapping from user behavior to underlying preferences.

\subsection{Further Analysis} 
\begin{wraptable}[11]{R}{0.55\linewidth}
\scriptsize
\centering
\caption{Evaluation of the quality of the final user hypotheses generated by \ours{}, by transferring them to non-finetuned base models.}
% \renewcommand{\arraystretch}{1.15}
% \resizebox{\columnwidth}{!}{%
\begin{tabular}{@{}l|cc|cc@{}}
\toprule
\multirow{2}{*}{Model}
  & \multicolumn{2}{c|}{Flight (Acc.)}
  & \multicolumn{2}{c}{HiCupid (Win Rate)} \\
\cmidrule(lr){2-3}\cmidrule(lr){4-5}
  & Base & w/ Refined Hyp.
    & Base & w/ Refined Hyp. \\
\midrule
Gemma4-4B
  & 40.9
  & $\mathbf{55.1}_{\textcolor{teal}{\scriptscriptstyle +14.2}}$
  & 50.0
  & $\mathbf{53.7}_{\textcolor{teal}{\scriptscriptstyle +3.7}}$ \\
Qwen3.5-4B
  & 40.3
  & $\mathbf{48.8}_{\textcolor{teal}{\scriptscriptstyle +8.5}}$
  & 50.0
  & $\mathbf{63.4}_{\textcolor{teal}{\scriptscriptstyle +13.4}}$ \\
Qwen3.5-9B
  & 41.7
  & $\mathbf{53.6}_{\textcolor{teal}{\scriptscriptstyle +11.9}}$
  & 50.0
  & $\mathbf{62.8}_{\textcolor{teal}{\scriptscriptstyle +12.8}}$ \\
\bottomrule
\end{tabular}%
% }

% \vspace{-5pt}
\label{tab:href-hyp-transfer}
\end{wraptable}

\paragraph{Reflective refinement produces reusable user hypotheses.}
Table~\ref{tab:href-hyp-transfer} presents our analysis on the quality of the user hypotheses learned by \ours{} through the hypothesis-transfer experiment. For each user, we take the final hypothesis generated by the \ours{} model and provide it to the corresponding non-finetuned base model for answer generation. During evaluation, each base model receives the hypothesis generated for the same user. The transferred hypotheses substantially improve performance across all models on both datasets, yielding average gains of 11.5 accuracy points on Flight Recommendation and 10.0 winrate points on HiCupid. These results indicate that our reflective refinement approach produces informative and reusable user hypotheses, suggesting that the resulting personalization benefits may extend beyond the model used to generate them.

\paragraph{Effect of reflection before consolidation.}

\begin{wraptable}[10]{R}{0.55\linewidth}
\centering
\scriptsize
\caption{Performance of \ours{} with and without reflection during refinement.}
\begin{tabular}{llcc}
\toprule
\textbf{Dataset} & \textbf{Method}
& \textbf{Gemma4-4B}
& \textbf{Qwen3.5-4B} \\
\midrule
\multirow{2}{*}{HiCupid}
& \ours{} w/o refl. & 72.8 & 83.8 \\
& \ours{}                            & \textbf{74.2} & \textbf{84.5} \\
\midrule
\multirow{2}{*}{Flight}
& \ours{} w/o refl. & 59.0 & 74.2 \\
& \ours{} & \textbf{62.9} & \textbf{76.0} \\
\bottomrule
\end{tabular}
\vspace{-10pt}
\label{tab:reflection-ablation}
\end{wraptable}
Table~\ref{tab:reflection-ablation} presents an ablation on HiCupid and Flight Recommendation to isolate the effect of the reflection step. For HiCupid, we evaluate users with long interaction histories. Our refinement step first reflects on the inferred hypothesis sets $\mathcal{I}$ and then consolidates them into an updated hypothesis set $H^*$ based on reflection. We compare direct consolidation of $\mathcal{I}$ without reflection.
Adding reflection before consolidation yields modest but consistent average gains of 1.0 and 2.8 points on HiCupid and Flight Recommendation, respectively, over direct consolidation of locally inferred hypothesis sets. Our qualitative analyses on both datasets in Appendix~\ref{app:qualitative-example} show that direct consolidation can retain broad, overlapping, or redundant preferences, whereas reflection before consolidation reorganizes the accumulated evidence into more specific and informative hypotheses. On Flight Recommendation, reflection also reevaluates recent behavioral evidence before consolidation, correcting misleading local inferences.

\paragraph{Hypothesis conditioning enables more focused personalization.}
We further present qualitative analyses between \textsc{SD} and \ours{} in Appendix~\ref{app:qualitative-sd-hypreflect}. On HiCupid, explicit hypotheses help \ours{} emphasize the most relevant preference (e.g., narrative-driven explanations), while \textsc{SD} remains more generic. On Flight Recommendation, they help prioritize the strongest discriminative preference (e.g., shorter duration) leading \ours{} to select the correct option. These examples suggest that shared hypothesis conditioning enables the student to better capture the user's style and preferences, leading to more targeted and personalized responses. This observation further suggests the shared abstraction layer may facilitate the distillation of personalized behaviors from the teacher to the student.

\subsection{Effect of Context Size and Reflection Window}

\begin{wrapfigure}[19]{R}{0.6\textwidth}
    \centering
    \scriptsize
    \vspace{-8pt}

    \includegraphics[width=0.49\linewidth]{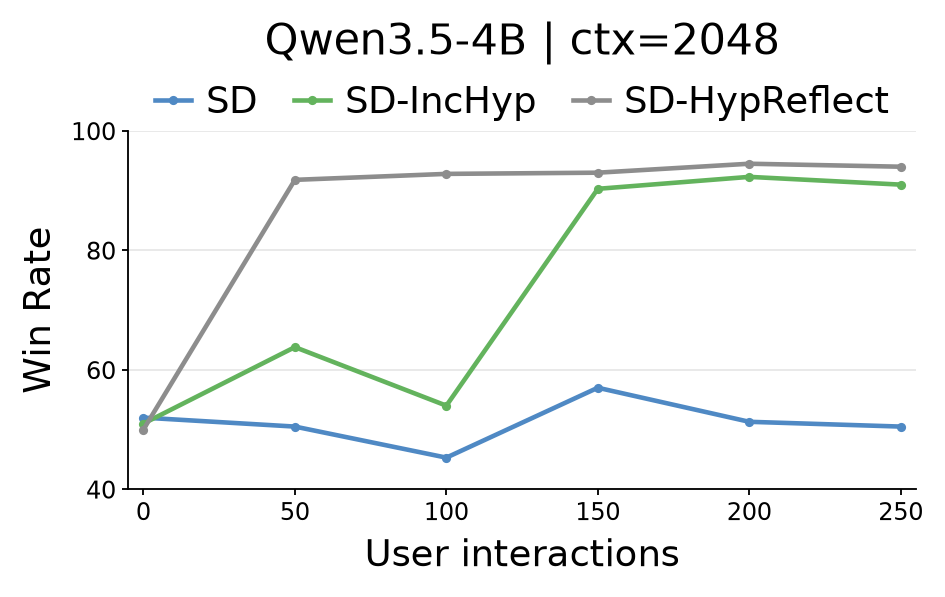}
    \hfill
    \includegraphics[width=0.49\linewidth]{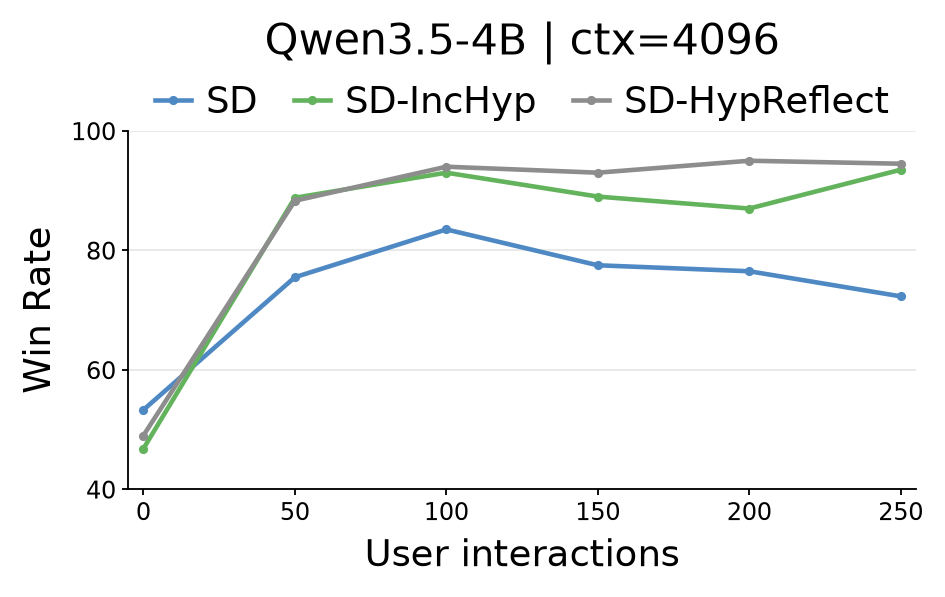}

    \vspace{2pt}

    \includegraphics[width=0.49\linewidth]{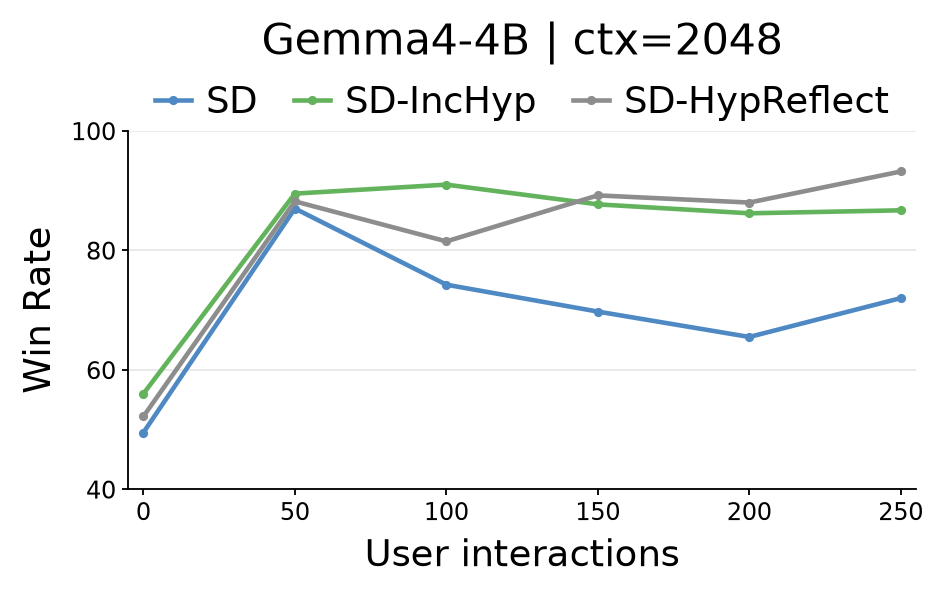}
    \hfill
    \includegraphics[width=0.49\linewidth]{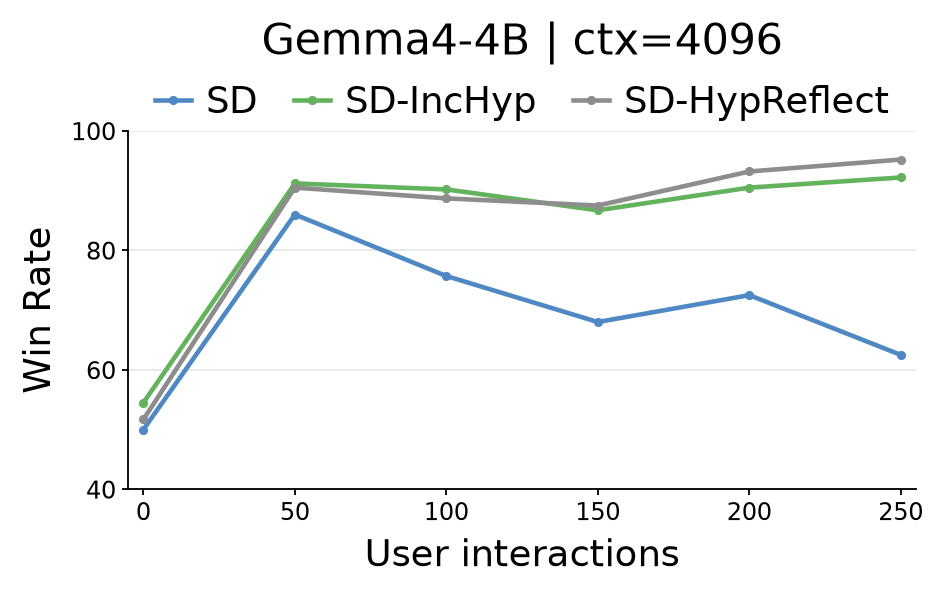}

    \caption{Win rate across user interactions under different context window sizes (2048, 4096) with Qwen3.5-4B and Gemma4-4B on HelpSteer2.} 
    \label{fig:context-window-comparison}
    \vspace{-10pt}
\end{wrapfigure}
% \end{wrapfigure}

\paragraph{Reflective refinement stabilizes online personalization across context budgets.}
Figure~\ref{fig:context-window-comparison} presents the effect of the context budget $B$ on HelpSteer2 in an online setting, where performance is evaluated as the model incrementally receives new evidence about the user. Since responses can contain up to 512 tokens, we evaluate context budgets of 2048 and 4096 tokens. \ours{} adapts quickly and remains stable under both budgets. With 2048 tokens, \textsc{SD} and \textsc{SD-IncHyp} show unstable early performance, especially for Qwen3.5-4B; for Gemma4-4B, both initially perform well but degrade as interaction history grows. Increasing the budget to 4096 improves \textsc{SD-IncHyp}, but it still underperforms \ours{}, while \textsc{SD} again degrades during later interactions. These results suggest that reflective refinement improves robustness by reconsidering evidence across interaction chunks, whereas retaining more tokens alone does not ensure better online personalization.

\begin{figure}[t]
    \centering
    \includegraphics[width=0.32\textwidth]{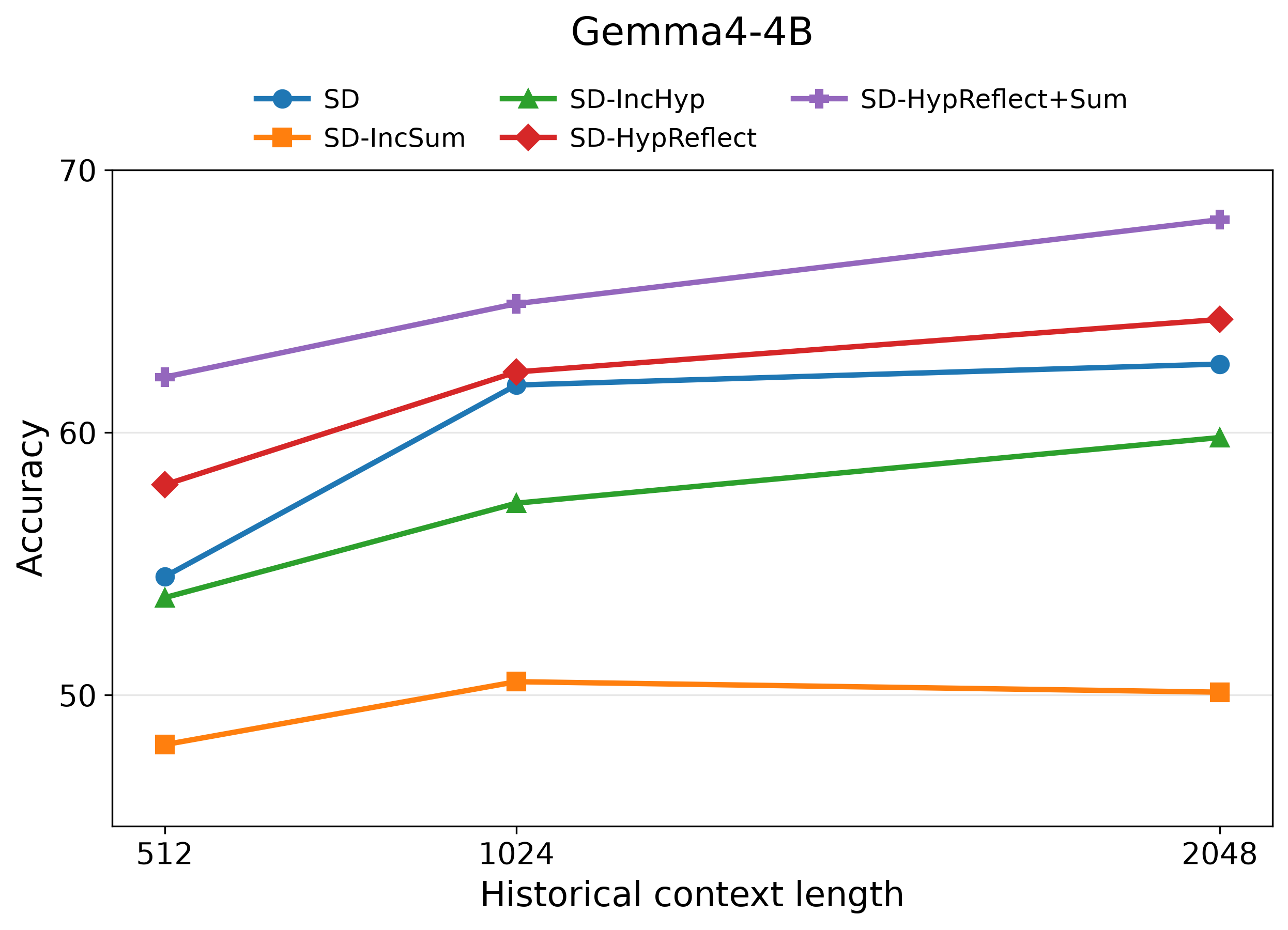}
    \hfill
    \includegraphics[width=0.32\textwidth]{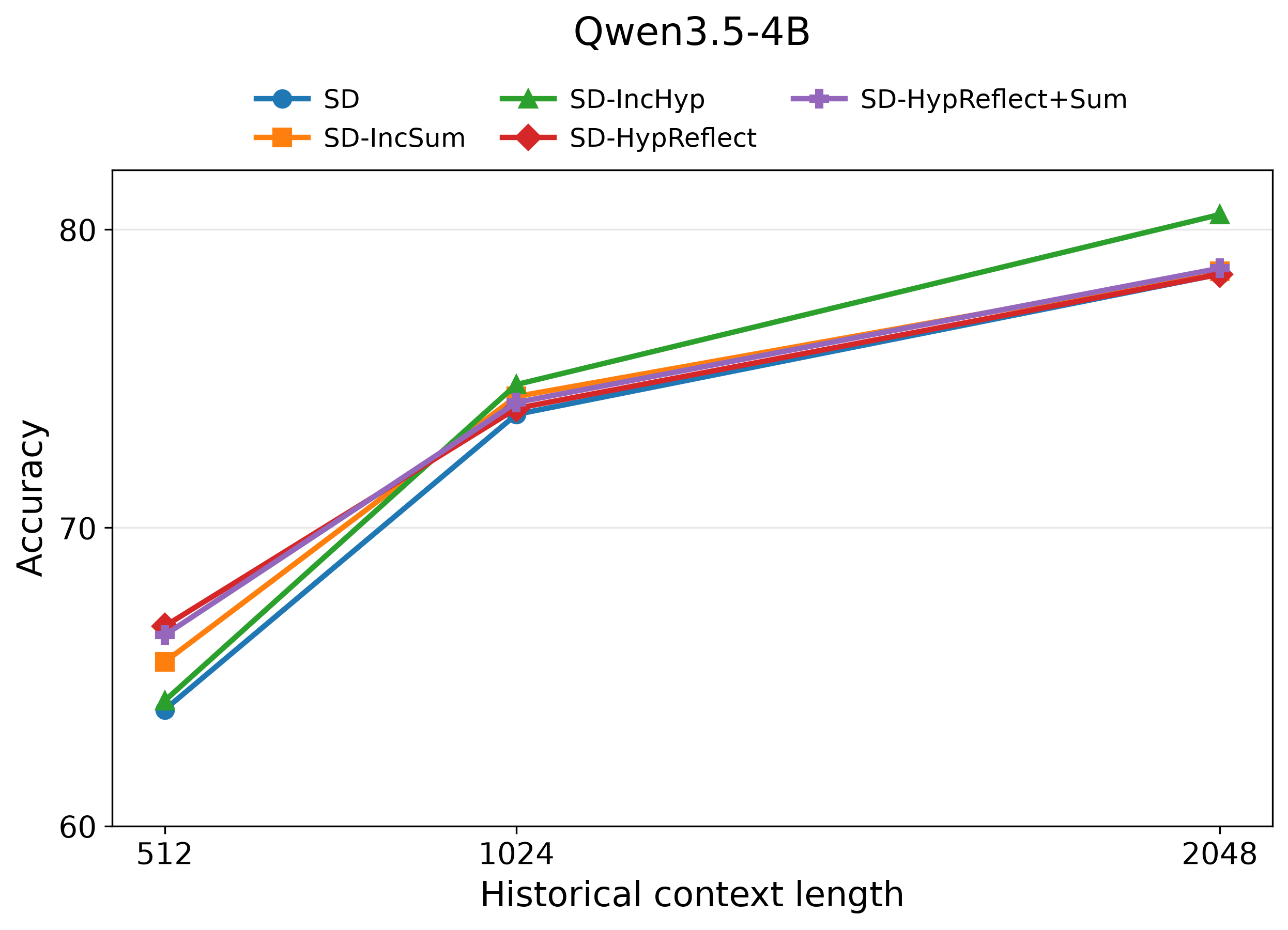}
    \hfill
    \includegraphics[width=0.32\textwidth]{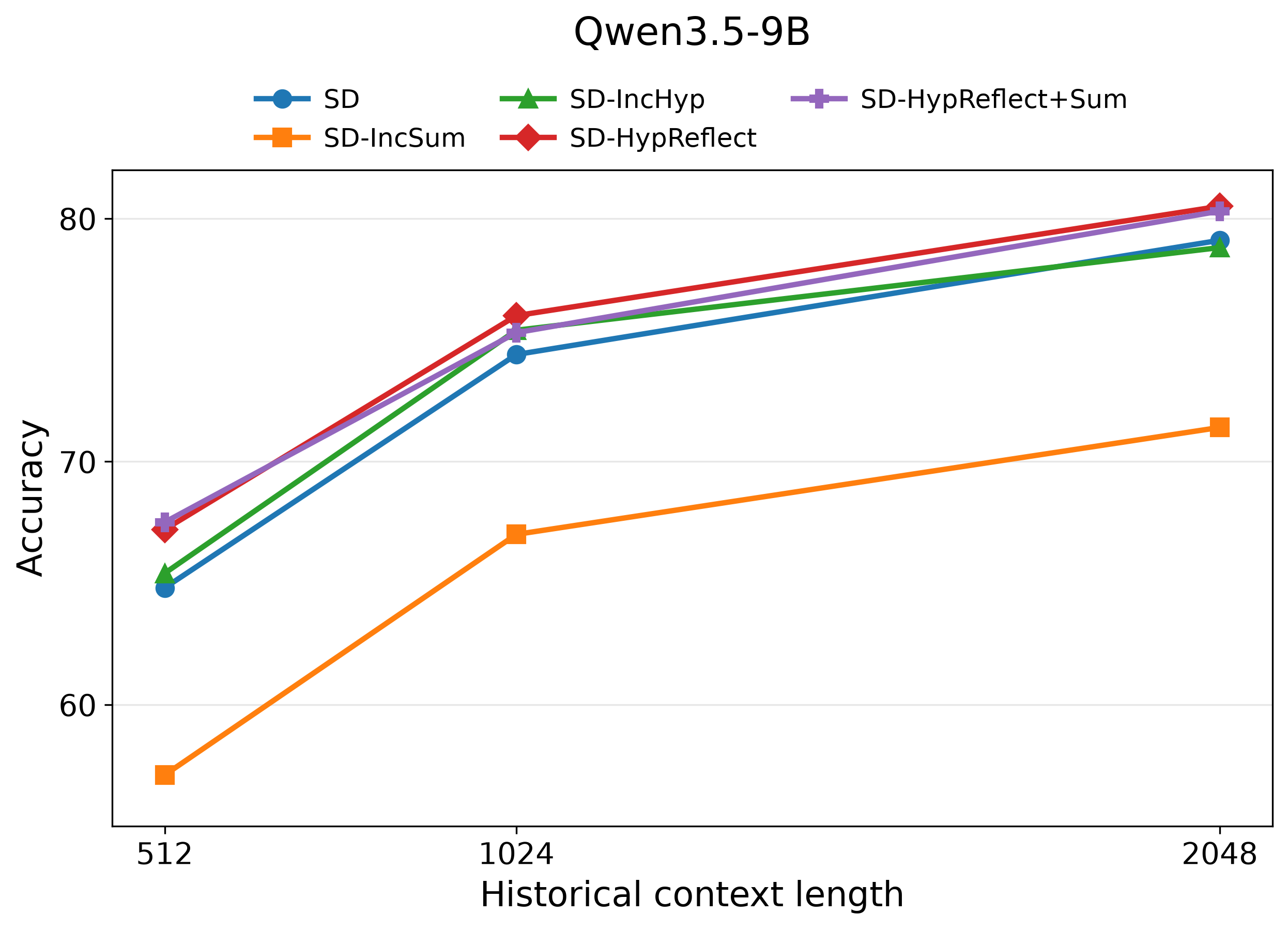}
    \caption{
        Accuracy across different context lengths for difference models on Flight.
    }
    \label{fig:context-length-results}
    % \vspace{-10pt}
\end{figure}

\paragraph{Preference inference from implicit behavior benefits as context budgets grow.}
Figure~\ref{fig:context-length-results} examines the effect of context budget on Flight Recommendation, where user preferences are inferred from implicit behavioral signals.
Flight uses fixed histories of short, structured choices over recurring attributes, so we evaluate smaller context budgets of 512, 1024, and 2048 tokens. Performance improves with larger context budgets, suggesting that additional context provides more evidence for inferring preferences from implicit behavioral signals. Across models and budgets, reflective refinement performs best in most settings, with clearer gains for Gemma4-4B and Qwen3.5-9B. Qwen3.5-4B, however, performs best with \textsc{SD-IncHyp} at larger budgets, suggesting that more context allows each user state to capture sufficient evidence, reducing the effects of noisy compression. These results demonstrate the robustness of reflective refinement across context budgets, supporting efficient continual personalization by reducing reliance on long, noisy interaction histories.

\paragraph{Effect of reflection window size.} 
Figure~\ref{fig:reflection-step} compares different reflection window sizes on a subset of HelpSteer2. The window controls how many previous hypothesis sets are considered during refinement. Smaller windows are unstable early on but improve as evidence accumulates. A window of 5 enables quick adaptation and maintains stability, while increasing it to 10 yields little additional gain. We use a window of 5 across datasets and find it performs consistently well, indicating a reasonable default across datasets.

\section{Related Work}
\label{sec:related_work}

\paragraph{Personalization in Conversational LLMs}
Personalized LLMs adapt responses as users' preferences, goals, and behaviors evolve. Prior work explores prompting, parameter adaptation, and alignment-based techniques~\citep{zhang2025personalizationlargelanguagemodels, liu2025surveypersonalizedlargelanguage, hwang-etal-2023-aligning}, while maintaining user representations through personas~\citep{zhang-etal-2026-personaagent, hwang-etal-2024-graph}, textual summaries~\citep{nam2026learningsummarizeuserinformation}, or RL-based profile updates~\citep{zhao2025teachinglanguagemodelsevolve}. However, these methods typically target specific personalization setting rather than continually integrating diverse user signals. \ours{} studies continual personalization across diverse user signals, including feedback-driven, multi-session, and behavioral settings.

\paragraph{User Modeling}
User modeling aims to infer and represent user preferences, goals, and behaviors for personalized interactions. Prior work models users through profiles, behavioral patterns, and predefined attributes~\citep{purificato2024usermodelinguserprofiling}. More recent approaches treat user understanding as an evolving inference process by maintaining beliefs over user goals~\citep{deng2026uncertaintyawareclarificationllmagents}, refining hypotheses through exploration-exploitation objectives~\citep{zhou-etal-2024-hypothesis}, inferring communication styles~\citep{garbacea2025hyperaligninterpretablepersonalizedllm}, explicitly
modeling mental-state hypotheses \citep{hwang-etal-2026-infusing}, and updating them through Bayesian inverse planning \citep{zhang2026autotomscalingmodelbasedmental}. However, these approaches often model specific aspects of user state rather than a unified, revisable representation integrating heterogeneous evidence over time.

\paragraph{Feedback-Driven Learning and Adaptation}
Recent work uses hindsight feedback to improve model behavior by distilling privileged information such as verified solutions~\citep{zhao2026selfdistilledreasoneronpolicyselfdistillation}, combining self-distillation with policy optimization and external feedback~\citep{hübotter2026reinforcementlearningselfdistillation}, or leveraging subsequent user responses for alignment~\citep{buening2026aligning}. Other methods adapt through interaction by refining user representations~\citep{wan2025enhancingpersonalizedmultiturndialogue, mehri2026multisessioncollablearninguserpreferences} or modeling uncertainty over user hypotheses~\citep{puri2026reachingmoderldistributional}. However, these approaches often rely on task-specific feedback or costly optimization, limiting scalable long-horizon personalization. \ours{} instead uses user signals as privileged supervision while maintaining an evolving preference representation across interactions.

\section{Conclusion}
We introduce \ours{}, a reliable, scalable framework for continual personalization that infers explicit, uncertainty-aware \emph{preference hypotheses} from diverse user signals, reflectively refines them over time, and incorporates them into the model via hypotheses-guided self-distillation. Our reflective refinement enables stable yet revisable user hypotheses over long interaction histories, providing a reliable user model that can guide self-distillation as new evidence accumulates.
Across three personalization settings, we show \ours{} consistently improves continual personalization approaches based on raw histories or incremental update, along with stability across context budgets.
Our results highlight the importance of maintaining explicit, revisable user hypotheses for achieving reliable and scalable continual personalization.

\section*{AI Use Statement}
% We used ChatGPT only for grammar checking and language refinement. The content, ideas, and code were developed and written by the authors.
We used ChatGPT solely for language editing and proofreading during the preparation of this manuscript. Specifically, it was used to identify grammatical errors, improve sentence structure, enhance clarity and readability, and refine wording and phrasing. ChatGPT was not used to generate, develop, or substantively modify the research ideas, methodology, experimental design, analysis, results, or conclusions. All scientific content, interpretations, claims, technical decisions, code, and experiments were developed, implemented, and verified by the authors. The authors reviewed and approved all AI-assisted edits and remain fully responsible for the final manuscript.

% \section*{Ethics Statement}
% \ej{people often ask syncophancy issue in personalization. Probably nice to discuss that issue here}
% Continual personalization systems rely on accumulating information from user interactions, which raises important privacy considerations. In real-world deployments, user data used to infer preferences should be collected with appropriate consent, handled securely, and subject to user control over storage, modification, and deletion. Although \ours{} maintains explicit preference hypotheses rather than directly exposing raw interaction histories, these representations may still encode sensitive information about users and should be treated as private user data.

% Personalization systems also introduce risks when inferred preferences are inaccurate, outdated, or overly generalized from limited evidence. Incorrect user models may lead to responses that reinforce assumptions about users or reduce their ability to explore alternative behaviors. Practical deployments of systems like \ours{} should incorporate mechanisms for preference correction and transparency to ensure that inferred preferences remain aligned with users' intended goals.

\section*{Ethics Statement}

Continual personalization systems rely on accumulating information from user interactions, which raises important privacy considerations. In real-world deployments, user data used to infer preferences should be collected with appropriate consent, handled securely, and subject to user control over storage, modification, and deletion. Although \ours{} maintains explicit preference hypotheses rather than directly exposing raw interaction histories, these representations may still encode sensitive information about users and should be treated as private user data.

Personalization systems also introduce risks when inferred preferences are inaccurate, outdated, or overly generalized from limited evidence. Incorrect user models may lead to responses that reinforce assumptions about users or reduce their ability to explore alternative behaviors. Practical deployments of systems like \ours{} should incorporate mechanisms for preference correction and transparency to ensure that inferred preferences remain aligned with users' intended goals.

Personalization may further amplify sycophancy, as models can interpret user preferences or prior feedback as signals to agree with the user's beliefs rather than merely adapt their responses. Over repeated interactions, this behavior may create a feedback loop in which the model increasingly reinforces the user's existing views, including inaccurate, biased, or harmful beliefs. This risk is particularly relevant when inferred preferences concern opinions or values rather than benign stylistic or task-related choices. Personalized systems should therefore distinguish between adapting to user preferences and endorsing user claims, preserve uncertainty when evidence is limited, and avoid treating agreement or positive feedback as evidence of factual correctness. They should also retain the ability to provide corrective information or alternative perspectives when appropriate.
\section*{Reproducibility Statement}
We are committed to ensuring the reproducibility of our results. Detailed descriptions of the experimental setup, including training, evaluation, and LLM-based judging procedures, as well as prompts, are provided in Appendix~\ref{appendix:exp-details} and in Appendix~\ref{app:href_prompts}. We plan to release our code upon acceptance.


\subsubsection*{Acknowledgments}
We thank Pouya Pezeshkpour, Farima Fatahi Bayat, Yanlin Feng, Seiji Maekawa, and other members of Megagon Labs for their insightful discussions and valuable feedback.

\bibliography{iclr2026_conference,anthology-1,anthology-2}
\bibliographystyle{iclr2026_conference}

\appendix
\label{sec:appendix}
\section{Limitations}

\paragraph{Modeling diverse and evolving user preferences}
\ours{} models user preferences by progressively refining evidence accumulated over interactions. In this work, we limit our experiments on three personalization settings, namely, explicit feedback, implicit behavioral signals, and conversational user profiles. However, users may express preferences through many other forms of interaction, including indirect corrections and contextual cues. Moreover, user preferences may change or evolve over time or depend on specific tasks and situations. Extending preference hypotheses to capture richer and more dynamic forms of user behavior remains an important direction.

\paragraph{Evaluation in real-world personalization settings.}
Our evaluation spans three complementary personalization datasets that capture distinct modes of user evidence for continual personalization. While this provides a broad assessment across different interaction paradigms, it may not fully reflect the complexity of long-term human interactions. Future work should complement these evaluations with longitudinal user studies and real-world deployments.

\paragraph{Computational cost of continual refinement.}
Maintaining explicit preference hypotheses introduces additional inference cost for hypothesis generation and reflective refinement. Although our bounded refinement strategy limits computation as interaction histories grow and avoids the cost of reward-model training and repeated policy optimization, continual personalization remains computationally demanding. Due to computational constraints, our experiments are limited to models up to 9B parameters. Exploring more efficient refinement strategies and evaluating stronger frontier models may lead to further improvements.

\paragraph{Preference hypothesis representation.} \ours{} assumes that user preferences can be represented as natural-language preference hypotheses with associated confidence scores. While this representation is interpretable, revisable, and directly usable for personalization, some preferences may be difficult to verbalize, interact in complex ways, or depend on latent contextual factors that are not easily captured through language alone. Future work could investigate richer representations that combine explicit preference hypotheses with structured, latent, or multimodal user representations.
\section{Methodology}
\label{appendix:methodology}

\subsection{Chunking}
\label{appendix:chunking}
In the offline setting, we partition the complete history into consecutive, non-overlapping chunks $C_1,\ldots,C_N$, each containing the longest sequence of complete interactions that fits within a token budget $B$. We then infer a set of local hypotheses from each chunk. In the online setting, interactions are accumulated sequentially until adding the next complete interaction would exceed $B$. At that point, local hypotheses are inferred from the completed chunk, and accumulation begins for the next chunk. Before the first chunk is completed, the model uses an empty hypothesis set; afterward, it uses the latest maintained hypotheses while accumulating interactions for the next set of local hypotheses.

\section{Experimental Setting Details}
\label{appendix:exp-details}
\subsection{Dataset Details}
\label{appendix:dataset}

\begin{figure*}
    \centering
    \includegraphics[width=\linewidth]{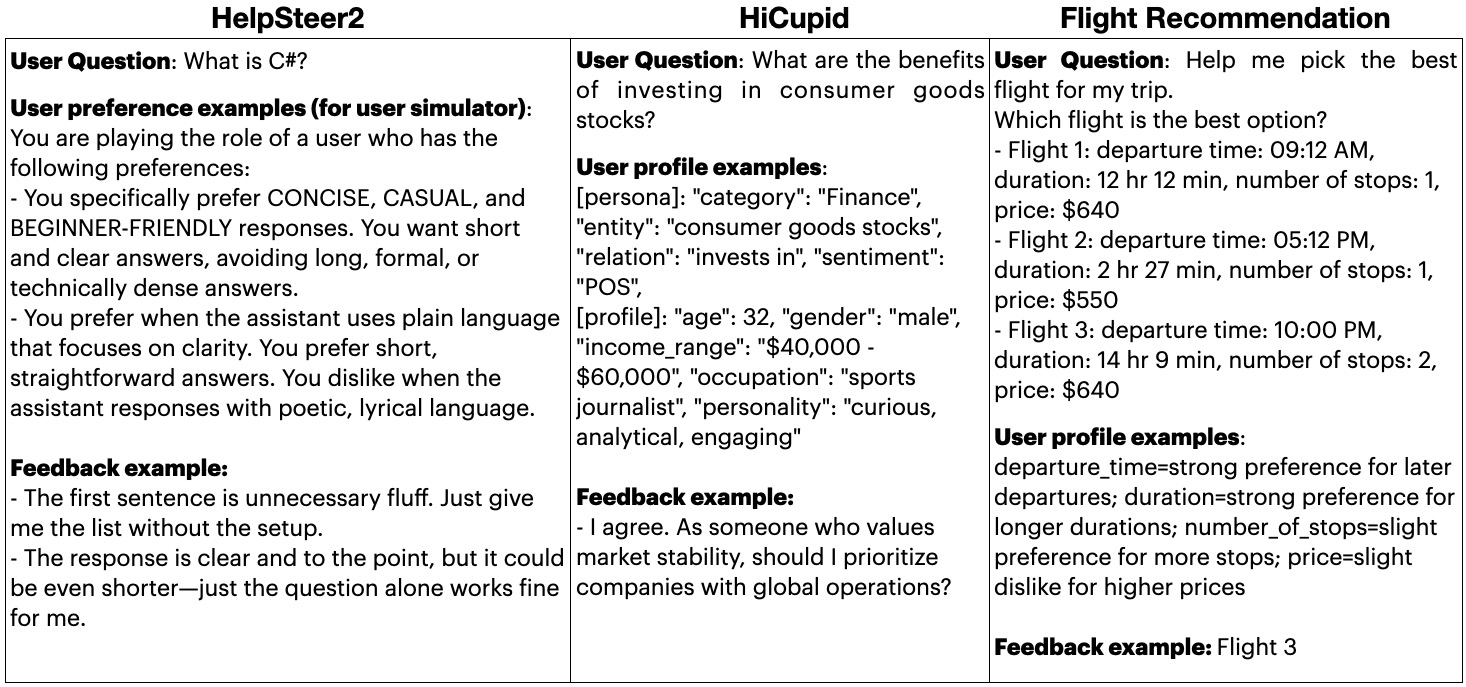}
    \caption{Example user question, preferences, and feedback (i.e., user's follow up message) of Helpsteer2, HiCupid, and Flight recommendation datasets.}
    \label{fig:dataset-example}
\end{figure*}

This section provides additional details about the construction of each evaluation benchmark. Figure~\ref{fig:dataset-example} presents an example of each dataset's user question, preferences, feedback (user's follow up question) of HelpSteer2, HiCupid, and Flight recommendation datasets.

\paragraph{HelpSteer2~\citep{wang2024helpsteer}.}
We use eight synthetic user profiles, where each profile is assigned a combination of three writing-style preferences adapted from TL;DR~\citep{cachola-etal-2020-tldr}. These preferences define the target response behavior for each user. 
At each interaction, the user simulator (\texttt{deepseek-v4-flash}) receives the generated response and provides feedback on one or two relevant preference dimensions. This creates a sparse feedback setting where the model must identify relevant preference signals over time. We train a separate model for each user profile.

\paragraph{HiCupid~\citep{mok-etal-2025-exploring}.}
HiCupid contains personalized question-answering examples associated with individual users and their historical interactions. 
We train a single model jointly on 9K QA examples from 300 users. We use user's profile data as a privileged information. For evaluation, we construct two settings: seen-user generalization, where the model receives previous interactions from users observed during training, and unseen-user generalization, where the model must personalize responses for users not encountered during training. The evaluation sets contain 1.8K prompts from seen users and 7.5K prompts from 250 held-out users.

\paragraph{Flight Recommendation~\citep{qiu2026bayesian}.}
In Flight data, each user trajectory contains a sequence of recommendation interactions where user selections provide indirect evidence about underlying preferences.
The benchmark contains 624 users and we use up to 25 interactions per user. User preferences may include non-obvious trade-offs, such as preferring higher prices or longer durations, requiring models to infer preferences from behavior rather than relying on standard assumptions. We evaluate across the original four-feature Flight setting, held-out feature settings with two to eight features, and transfer to the hotel recommendation domain.

\subsection{Training Hyperparameters}
\label{app:train-details}
We initialize our configurations based on the implementation of \citet{buening2026aligning} and the corresponding dataset-specific codebases, and tune the learning rate and batch size for each dataset. We explore learning rates of \{$1\times10^{-6}, 5\times10^{-6}, 1\times10^{-5}$\} and observe relatively consistent performance on HelpSteer and HiCupid, whereas Flight performs best with $1\times10^{-5}$. We also evaluate batch sizes of 2, 4, and 8 on Flight and 16 and 32 on HiCupid. HiCupid is relatively insensitive to batch size, while Flight benefits from smaller batches; we therefore use batch sizes of 32 and 4, respectively. Performance is otherwise stable across the tested hyperparameters, except for the maximum history/chunk and generation lengths, which are limited by GPU memory. When memory permits, larger history/chunk sizes generally improve performance by preserving more information from previous interactions. The final configurations are reported in Table~\ref{tab:training_hyperparameters}.
We used NVIDIA A800 GPUs, and each experiment can be run on a single GPU node. Training and evaluation take approximately 15 hours on HelpSteer2, up to 17 hours on Flight, and up to 10 hours on HiCupid.

\begin{table}[t]
\centering
\scriptsize
\caption{Training hyperparameters used for each dataset.}
\begin{tabular}{lccc}
\toprule
\textbf{Hyperparameter}
& \textbf{HelpSteer}
& \textbf{Flight}
& \textbf{HiCupid} \\
\midrule
Number of epochs
    & Online
    & 1
    & 1 \\
Batch size
    & 1
    & 4
    & 32 \\
Learning rate
    & $1\times10^{-6}$
    & $1\times10^{-5}$
    & $1\times10^{-6}$ \\
LoRA rank
    & 32
    & 32
    & 32 \\
LoRA alpha
    & 64
    & 64
    & 64 \\
LoRA dropout
    & 0
    & 0
    & 0 \\
History/chunk size
    & 4,096
    & 1,024
    & 1,024 \\
Max generation length
    & 512
    & 400
    & 512 \\
Temperature
    & 0.7
    & 0.7
    & 0.7 \\
Top-$p$
    & 1.0
    & 1.0
    & 1.0 \\
Top-$k$
    & 20
    & 64
    & 64 \\
Reflection window size
&5&5&5 \\
\bottomrule
\end{tabular}
\label{tab:training_hyperparameters}
\end{table}

\subsection{Evaluation Details}
\label{app:eval-detail}
We follow the original dataset papers for evaluation, including an LLM-based user simulator and pairwise LLM judging where applicable. We use DeepSeek-V4-Flash as the judge for HelpSteer and HiCupid, while Flight is evaluated through exact matching against the ground-truth choices. We additionally compare the judgments of DeepSeek-V4-Flash with those of DeepSeek-V4-Pro and Qwen3.7-Plus (see Table~\ref{tab:llm-judge-validity} and Appendix~\ref{app:llm-judge-validity}) and observe consistent evaluation outcomes across the models. Given its comparable evaluation performance and lower inference cost, we select DeepSeek-V4-Flash as a practical balance between evaluation quality and cost. The complete evaluation settings are reported in Table~\ref{tab:evaluation_settings}.
See \citet{buening2026aligning} (\href{https://github.com/lasgroup/user_interactions}{repo}) for the LLM judge and user simulator prompts used for HelpSteer2, and \citet{mok-etal-2025-exploring} (\href{https://github.com/12kimih/hicupid}{repo}) for details of the LLM judge prompt used for HiCupid. Our code is also built based on these two repos.
Running DeepSeek-V4-Flash as both the LLM judge and user simulator on HelpSteer2 costs approximately 10 USD and using it as the LLM judge on HiCupid costs approximately 10 USD.

\begin{table}[t]
\centering
\scriptsize
\caption{Evaluation settings used for each dataset.}
\begin{tabular}{lccc}
\toprule
\textbf{Evaluation detail}
& \textbf{HelpSteer}
& \textbf{Flight}
& \textbf{HiCupid} \\
\midrule
Evaluation data
    & 256 validation prompts
    & 624 users / 20 questions per user
    & 300 Seen users, 300 Unseen users \\
Evaluation frequency
    & Every 50 interactions
    & After training
    & After training \\
Evaluator
    & deepseek-v4-flash
    & Exact answer matching
    & deepseek-v4-flash \\
Primary metric
    & Pairwise win rate
    & Accuracy
    & Pairwise win rate \\
Reference
    & Frozen base model
    & Ground-truth Flight
    & Frozen base model \\
\bottomrule
\end{tabular}
\label{tab:evaluation_settings}
\end{table}

\section{LLM Judge Validity}
\label{app:llm-judge-validity}

We validate the use of \texttt{deepseek-v4-flash} as our primary LLM judge by comparing its evaluations with those of \texttt{deepseek-v4-pro} and \texttt{qwen3.7-plus}. Although the absolute scores vary across judges, all three produce the same ranking: \ours{} consistently outperforms \textsc{SD-IncHyp}, which in turn outperforms \textsc{SD}. This agreement indicates that our main conclusions are robust to the choice of LLM judge.

\begin{wraptable}[11]{r}{0.5\linewidth}
\centering
\caption{Evaluation results using different LLM judges on the subset of HelpSteer2. All judges produce the same relative ranking across methods.}
\scalebox{0.8}{
\begin{tabular}{@{}lccc@{}}
\toprule
& \multicolumn{1}{l}{SD} & \multicolumn{1}{l}{SD-IncHyp} & \multicolumn{1}{l}{\ours{}} \\
\midrule
Deepseek-v4-flash & 77.1 & 90.3 & 93.0 \\
Deepseek-v4-pro   & 74.1 & 86.3 & 87.1 \\
Qwen3.7-plus      & 71.8 & 83.4 & 86.4 \\
\bottomrule
\end{tabular}
}
\label{tab:llm-judge-validity}
\end{wraptable}

\section{Additional results}

\subsection{Qualitative Examples for Reflection}
\label{app:qualitative-example}
\begin{figure*}[t]
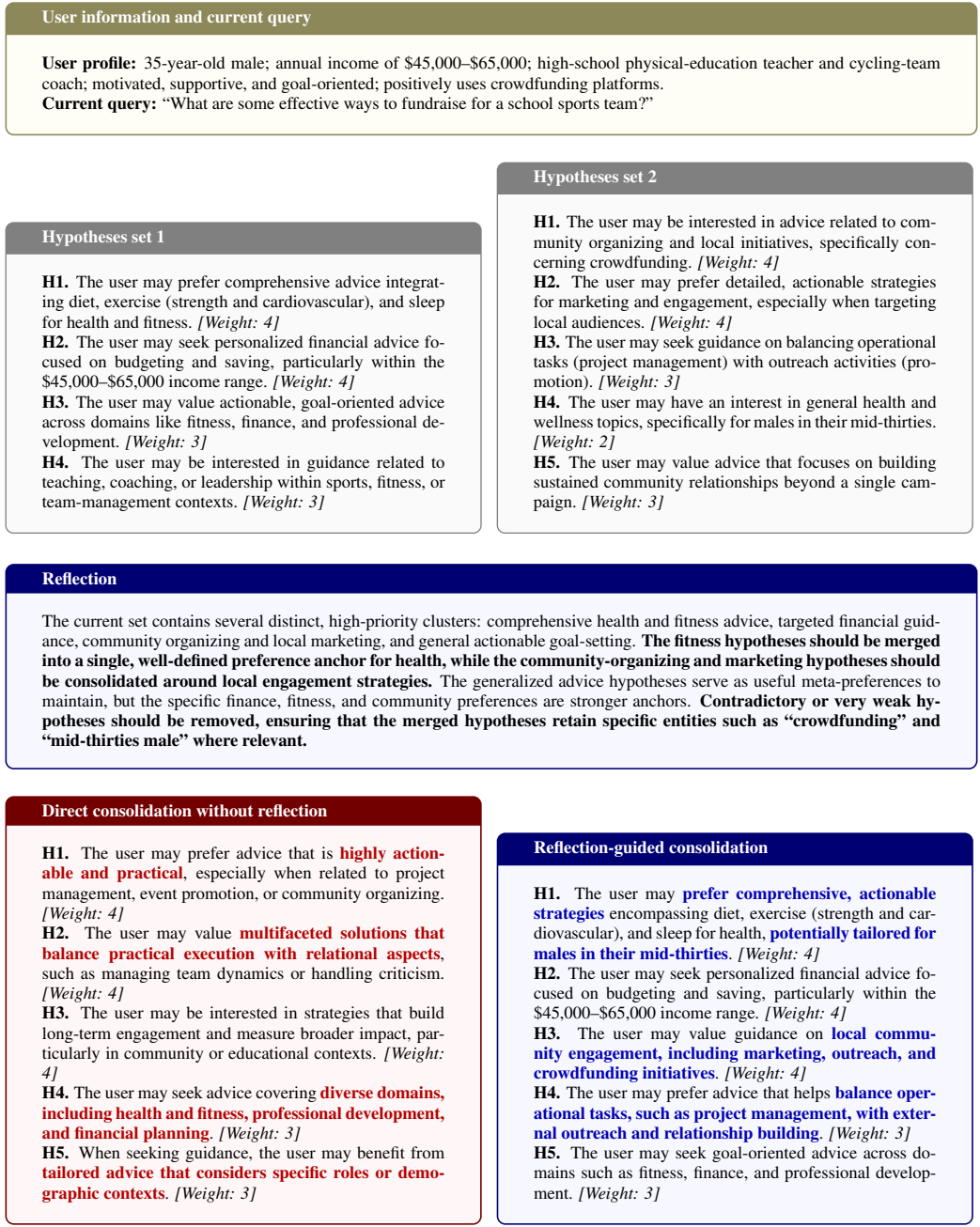

\centering
\begin{minipage}{0.98\textwidth}
\scriptsize
\begin{tcolorbox}[title=\textbf{User information and current query}, colback=yellow!4, colframe=yellow!45!black, boxrule=0.6pt, arc=1mm]
\textbf{User profile:} 35-year-old male; annual income of \$45,000--\$65,000; high-school physical-education teacher and cycling-team coach; motivated, supportive, and goal-oriented; positively uses crowdfunding platforms.
\\
\textbf{Current query:} ``What are some effective ways to fundraise for a school sports team?''
\end{tcolorbox}
\vspace{2mm}
\begin{minipage}[t]{0.49\textwidth}
\begin{tcolorbox}[title=\textbf{Hypotheses set 1}, colback=gray!4, colframe=black!50, boxrule=0.5pt, arc=1mm]
\textbf{H1.} The user may prefer comprehensive advice integrating diet, exercise (strength and cardiovascular), and sleep for health and fitness. \textit{[Weight: 4]}\\
\textbf{H2.} The user may seek personalized financial advice focused on budgeting and saving, particularly within the \$45,000--\$65,000 income range. \textit{[Weight: 4]}\\
\textbf{H3.} The user may value actionable, goal-oriented advice across domains like fitness, finance, and professional development. \textit{[Weight: 3]}\\
\textbf{H4.} The user may be interested in guidance related to teaching, coaching, or leadership within sports, fitness, or team-management contexts. \textit{[Weight: 3]}
\end{tcolorbox}
\end{minipage}
\hfill
\begin{minipage}[t]{0.49\textwidth}
\begin{tcolorbox}[title=\textbf{Hypotheses set 2}, colback=gray!4, colframe=black!50, boxrule=0.5pt, arc=1mm]
\textbf{H1.} The user may be interested in advice related to community organizing and local initiatives, specifically concerning crowdfunding. \textit{[Weight: 4]}\\
\textbf{H2.} The user may prefer detailed, actionable strategies for marketing and engagement, especially when targeting local audiences. \textit{[Weight: 4]}\\
\textbf{H3.} The user may seek guidance on balancing operational tasks (project management) with outreach activities (promotion). \textit{[Weight: 3]}\\
\textbf{H4.} The user may have an interest in general health and wellness topics, specifically for males in their mid-thirties. \textit{[Weight: 2]}\\
\textbf{H5.} The user may value advice that focuses on building sustained community relationships beyond a single campaign. \textit{[Weight: 3]}
\end{tcolorbox}
\end{minipage}
\vspace{2mm}
\begin{tcolorbox}[title=\textbf{Reflection}, colback=blue!3, colframe=blue!45!black, boxrule=0.6pt, arc=1mm]
The current set contains several distinct, high-priority clusters: comprehensive health and fitness advice, targeted financial guidance, community organizing and local marketing, and general actionable goal-setting. \textbf{The fitness hypotheses should be merged into a single, well-defined preference anchor for health, while the community-organizing and marketing hypotheses should be consolidated around local engagement strategies.} The generalized advice hypotheses serve as useful meta-preferences to maintain, but the specific finance, fitness, and community preferences are stronger anchors. \textbf{Contradictory or very weak hypotheses should be removed, ensuring that the merged hypotheses retain specific entities such as ``crowdfunding'' and ``mid-thirties male'' where relevant.}
\end{tcolorbox}
\vspace{2mm}
\begin{minipage}[t]{0.49\textwidth}
\begin{tcolorbox}[
title=\textbf{Direct consolidation without reflection},
colback=red!3,
colframe=red!45!black,
boxrule=0.6pt,
arc=1mm
]
\textbf{H1.} The user may prefer advice that is
\refinementissue{highly actionable and practical},
especially when related to project management, event promotion, or community organizing.
\textit{[Weight: 4]}\\
\textbf{H2.} The user may value
\refinementissue{multifaceted solutions that balance practical execution with relational aspects},
such as managing team dynamics or handling criticism.
\textit{[Weight: 4]}\\
\textbf{H3.} The user may be interested in strategies that build long-term engagement and measure broader impact, particularly in community or educational contexts.
\textit{[Weight: 4]}\\
\textbf{H4.} The user may seek advice covering
\refinementissue{diverse domains, including health and fitness, professional development, and financial planning}.
\textit{[Weight: 3]}\\
\textbf{H5.} When seeking guidance, the user may benefit from
\refinementissue{tailored advice that considers specific roles or demographic contexts}.
\textit{[Weight: 3]}
\end{tcolorbox}
\end{minipage}
\hfill
\begin{minipage}[t]{0.49\textwidth}
\begin{tcolorbox}[
title=\textbf{Reflection-guided consolidation},
colback=blue!3,
colframe=blue!45!black,
boxrule=0.6pt,
arc=1mm
]
\textbf{H1.} The user may \reflectionhelp{prefer comprehensive, actionable strategies} encompassing diet, exercise (strength and cardiovascular), and sleep for health, \reflectionhelp{potentially tailored for males in their mid-thirties}.
\textit{[Weight: 4]}\\
\textbf{H2.} The user may seek personalized financial advice focused on budgeting and saving, particularly within the \$45,000--\$65,000 income range.
\textit{[Weight: 4]}\\
\textbf{H3.} The user may value guidance on
\reflectionhelp{local community engagement, including marketing, outreach, and crowdfunding initiatives}.
\textit{[Weight: 4]}\\
\textbf{H4.} The user may prefer advice that helps
\reflectionhelp{balance operational tasks, such as project management, with external outreach and relationship building}.
\textit{[Weight: 3]}\\
\textbf{H5.} The user may seek goal-oriented advice across domains such as fitness, finance, and professional development.
\textit{[Weight: 3]}
\end{tcolorbox}
\end{minipage}
\vspace{1mm}
\end{minipage}
\caption{An example of hypothesis consolidation with and without explicit reflection.}
\label{fig:reflection-hypothesis-refinement1}
\end{figure*}

\begin{figure*}[t]
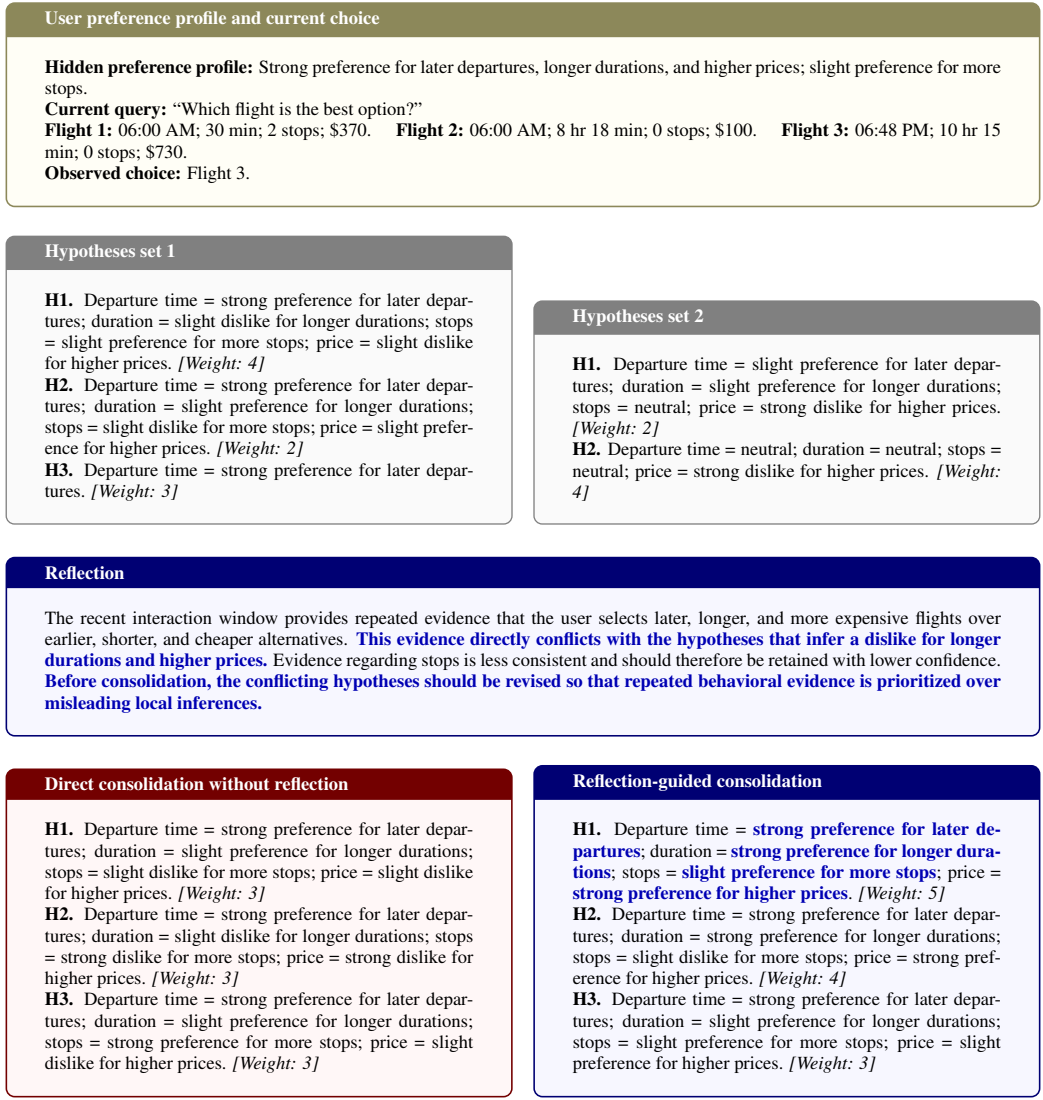

\centering
\begin{minipage}{0.98\textwidth}
\scriptsize

\begin{tcolorbox}[
title=\textbf{User preference profile and current choice},
colback=yellow!4,
colframe=yellow!45!black,
boxrule=0.6pt,
arc=1mm
]
\textbf{Hidden preference profile:}
Strong preference for later departures, longer durations, and higher prices;
slight preference for more stops.
\\
\textbf{Current query:} ``Which flight is the best option?''
\\
\textbf{Flight 1:} 06:00 AM; 30 min; 2 stops; \$370.
\quad
\textbf{Flight 2:} 06:00 AM; 8 hr 18 min; 0 stops; \$100.
\quad
\textbf{Flight 3:} 06:48 PM; 10 hr 15 min; 0 stops; \$730.
\\
\textbf{Observed choice:} Flight 3.
\end{tcolorbox}

\vspace{2mm}

\begin{minipage}[t]{0.49\textwidth}
\begin{tcolorbox}[
title=\textbf{Hypotheses set 1},
colback=gray!4,
colframe=black!50,
boxrule=0.5pt,
arc=1mm
]
\textbf{H1.}
Departure time = strong preference for later departures;
duration = slight dislike for longer durations;
stops = slight preference for more stops;
price = slight dislike for higher prices.
\textit{[Weight: 4]}
\\
\textbf{H2.}
Departure time = strong preference for later departures;
duration = slight preference for longer durations;
stops = slight dislike for more stops;
price = slight preference for higher prices.
\textit{[Weight: 2]}
\\
\textbf{H3.}
Departure time = strong preference for later departures.
\textit{[Weight: 3]}
\end{tcolorbox}
\end{minipage}
\hfill
\begin{minipage}[t]{0.49\textwidth}
\begin{tcolorbox}[
title=\textbf{Hypotheses set 2},
colback=gray!4,
colframe=black!50,
boxrule=0.5pt,
arc=1mm
]
\textbf{H1.}
Departure time = slight preference for later departures;
duration = slight preference for longer durations;
stops = neutral;
price = strong dislike for higher prices.
\textit{[Weight: 2]}
\\
\textbf{H2.}
Departure time = neutral;
duration = neutral;
stops = neutral;
price = strong dislike for higher prices.
\textit{[Weight: 4]}
\end{tcolorbox}
\end{minipage}

\vspace{2mm}

\begin{tcolorbox}[
title=\textbf{Reflection},
colback=blue!3,
colframe=blue!45!black,
boxrule=0.6pt,
arc=1mm
]
The recent interaction window provides repeated evidence that the user selects later, longer, and more expensive flights over earlier, shorter, and cheaper alternatives.
\reflectionhelp{This evidence directly conflicts with the hypotheses that infer a dislike for longer durations and higher prices.}
Evidence regarding stops is less consistent and should therefore be retained with lower confidence.
\reflectionhelp{Before consolidation, the conflicting hypotheses should be revised so that repeated behavioral evidence is prioritized over misleading local inferences.}
\end{tcolorbox}

\vspace{2mm}

\begin{minipage}[t]{0.49\textwidth}
\begin{tcolorbox}[
title=\textbf{Direct consolidation without reflection},
colback=red!3,
colframe=red!45!black,
boxrule=0.6pt,
arc=1mm
]
\textbf{H1.}
Departure time = strong preference for later departures;
duration = slight preference for longer durations;
stops = slight dislike for more stops;
price = slight dislike for higher prices.
\textit{[Weight: 3]}
\\
\textbf{H2.}
Departure time = strong preference for later departures;
duration = slight dislike for longer durations;
stops = strong dislike for more stops;
price = strong dislike for higher prices.
\textit{[Weight: 3]}
\\
\textbf{H3.}
Departure time = strong preference for later departures;
duration = slight preference for longer durations;
stops = strong preference for more stops;
price = slight dislike for higher prices.
\textit{[Weight: 3]}
\end{tcolorbox}
\end{minipage}
\hfill
\begin{minipage}[t]{0.49\textwidth}
\begin{tcolorbox}[
title=\textbf{Reflection-guided consolidation},
colback=blue!3,
colframe=blue!45!black,
boxrule=0.6pt,
arc=1mm
]
\textbf{H1.}
Departure time =
\reflectionhelp{strong preference for later departures};
duration =
\reflectionhelp{strong preference for longer durations};
stops =
\reflectionhelp{slight preference for more stops};
price =
\reflectionhelp{strong preference for higher prices}.
\textit{[Weight: 5]}
\\
\textbf{H2.}
Departure time = strong preference for later departures;
duration = strong preference for longer durations;
stops = slight dislike for more stops;
price = strong preference for higher prices.
\textit{[Weight: 4]}
\\
\textbf{H3.}
Departure time = strong preference for later departures;
duration = slight preference for longer durations;
stops = slight preference for more stops;
price = slight preference for higher prices.
\textit{[Weight: 3]}
\end{tcolorbox}
\end{minipage}

\vspace{1mm}

\end{minipage}

\caption{
A qualitative Flight Recommendation example comparing direct and reflection-guided hypothesis consolidation.
Direct consolidation propagates misleading inferences about duration and price, whereas reflection re-evaluates conflicting evidence across the recent interaction window and produces a top hypothesis that exactly matches the hidden preference profile.
}
\label{fig:reflection-hypothesis-refinement2}
\end{figure*}

% \subsection{How does reflection improve hypothesis consolidation?}
Figures~\ref{fig:reflection-hypothesis-refinement1} and \ref{fig:reflection-hypothesis-refinement2} presents some qualitative examples that compares between hypotheses consolidation without reflection and with reflection.
We find that direct consolidation tends to transform local hypotheses into broad and partially overlapping meta-preferences, such as favoring actionable, multifaceted, or tailored advice. These abstractions omit concrete concepts from the original hypothesis sets and provide less distinctive guidance for personalization. Reflection instead explicitly identifies which hypotheses should be merged, retained, or removed, preserving specific preference anchors such as crowdfunding, local outreach, marketing, and the balance between project management and relationship building. This example suggests that reflection reduces vague generalization and redundancy while producing more specific and informative preference hypotheses.

% \noindent\textbf{Interpretation.} The selected flight departs later, has the longest duration among the relevant alternatives, and costs \$730 rather than \$100. Nevertheless, the latest local hypotheses infer a dislike for higher prices, and direct consolidation retains this error. Reflection explicitly identifies recent choices that contradict the negative duration and price preferences, overrides those assignments, and recovers the exact hidden preference profile.

\subsection{Qualitative Examples of Final Responses: SD vs. \ours{}}
\label{app:qualitative-sd-hypreflect}

Figures~\ref{fig:sd-vs-href1} and~\ref{fig:sd-vs-href2} present qualitative comparisons between the final responses generated by SD and \ours{}, together with the preference hypotheses used by \ours{}. In the HiCupid example, both methods provide generally relevant advice, but \ours{} more explicitly emphasizes narrative structure and personification, reflecting the highest-weight hypothesis inferred from the user's interest in engaging, science-oriented content. In the Flight example, the hypotheses help \ours{} resolve competing preferences by prioritizing the user's strong dislike of long travel durations. Because all three flights have the same price, \ours{} correctly selects Flight~2, which has the shortest duration, whereas SD selects Flight~3. These examples illustrate how explicit, weighted hypotheses can help the model identify the most relevant user preferences and prioritize them when generating personalized responses.

\begin{figure*}[p]
\centering
\begin{comparisonbox}
\footnotesize
\medskip
\textbf{Dataset:} HiCupid \\
\textbf{Hidden user profile.}
32-year-old man; sports journalist; income \$40,000--\$60,000;
curious, analytical, and engaging; follows Mark Rober.

\medskip
\textbf{User query.}
What are some engaging ways to explain complex topics to a general audience?

\medskip
\textbf{Hypotheses supplied to \ours{}.} \\
\textbf{H1:} The user may prefer narrative-driven educational content that blends complex scientific concepts with relatable storytelling and personification (weight 5). \\
\textbf{H2:} The user may value high production quality while remaining open to cost-effective or free content-creation tools (weight 4). \\
\textbf{H3:} The user may prefer career guidance that balances existing skills with advanced opportunities in sports journalism (weight 4). \\
\textbf{H4:} The user may prefer learning methods that combine theoretical knowledge with hands-on application (weight 4). \\
\textbf{H5:} The user may prefer actionable health or financial guidance tailored to men in their early thirties (weight 3).

\medskip
\textbf{Final answer with \ours{}.}
To explain complex topics to a general audience, lean heavily into analogies and metaphors, as these create instant relatable connections.
\textcolor{blue}{Frame the topic as a narrative, personifying concepts or structuring the explanation like a mystery to be solved.}
Furthermore, focus on the ``why it matters''---connecting the science directly to the audience's daily life or major societal issues makes the information immediately relevant and engaging.

\medskip
\textbf{Final answer with SD.}
To explain complex topics to a general audience, employ strong analogies and metaphors that connect the abstract concept to everyday experiences.
Visual demonstrations and simplified models are crucial for making the invisible visible.
Additionally, framing the topic as a compelling story---highlighting the ``why'' and the real-world implications---keeps the audience hooked.
\end{comparisonbox}
\caption{Final response comparison between SD and \ours{} on HiCupid}
\label{fig:sd-vs-href1}
\end{figure*}

\begin{figure*}[p]
\centering
\begin{comparisonbox}
\footnotesize
\textbf{Dataset:} Flight Recommendation \hfill \\
\medskip
\textbf{Hidden user profile.}
Slight preference for later departures; strong dislike for longer durations;
slight preference for more stops; strong dislike for higher prices.

\medskip
\textbf{User query.}
Which flight is the best option? \\
- Flight 1: 10:00 PM; 20 hr; 0 stops; \$910. \\
- Flight 2: 07:36 AM; 8 hr 18 min; 2 stops; \$910. \\
- Flight 3: 09:12 AM; 14 hr 9 min; 2 stops; \$910. \\

\medskip
\textbf{Hypotheses supplied to \ours{}.} \\
\textbf{H1:} Slight preference for later departures; strong dislike for longer durations; slight preference for more stops; strong dislike for higher prices (weight 4). \\
\textbf{H2:} Strong dislike for later departures, longer durations, more stops, and higher prices (weight 3). \\
\textbf{H3:} Slight preference for later departures; strong dislike for longer durations, more stops, and higher prices (weight 3). \\
\medskip

\textbf{User's choice:} Flight 2. \\
\textbf{\ours{} answer:} \textcolor{blue}{Flight 2 (correct).} \\
\textbf{\textbf{SD} answer:} \textcolor{red}{Flight 3 (incorrect).}
\end{comparisonbox}
\caption{Final response comparison between SD and \ours{} on Flight Recommendation}
\label{fig:sd-vs-href2}
\end{figure*}

% \textbf{Observed effect.}
% The highest-weight hypothesis exactly recovers the four-dimensional profile.
% Because all prices are equal, it helps prioritize the much shorter duration and preferred number of stops.

% Href more directly realizes the narrative-driven and personification hypothesis; A gives a broader, more conventional explanation.
\begin{figure}
    \centering
    \includegraphics[width=0.4\linewidth]{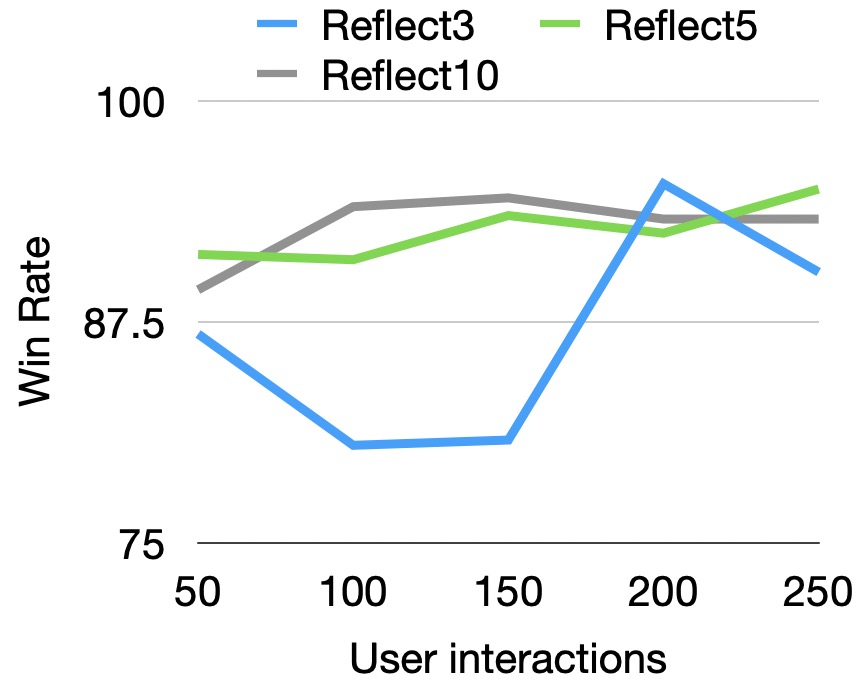}
    \caption{Performance with \ours{} across different reflection window sizes on HelpSteer2}
    \label{fig:reflection-step}
% \end{wrapfigure}
\end{figure}
\subsection{Reflection window size variation}
\label{app:reflection-window}
Figure~\ref{fig:reflection-step} compares performance under different reflection window sizes on a subset of the HelpSteer2 setting at every 50 interactions.
% \begin{figure}

% \section{AI Usage for Writing}
% We used ChatGPT only for grammar checking and language refinement. The content, ideas, and code were developed and written by the authors.

\section{Prompts}
\label{app:href_prompts}

\begin{figure*}[t]
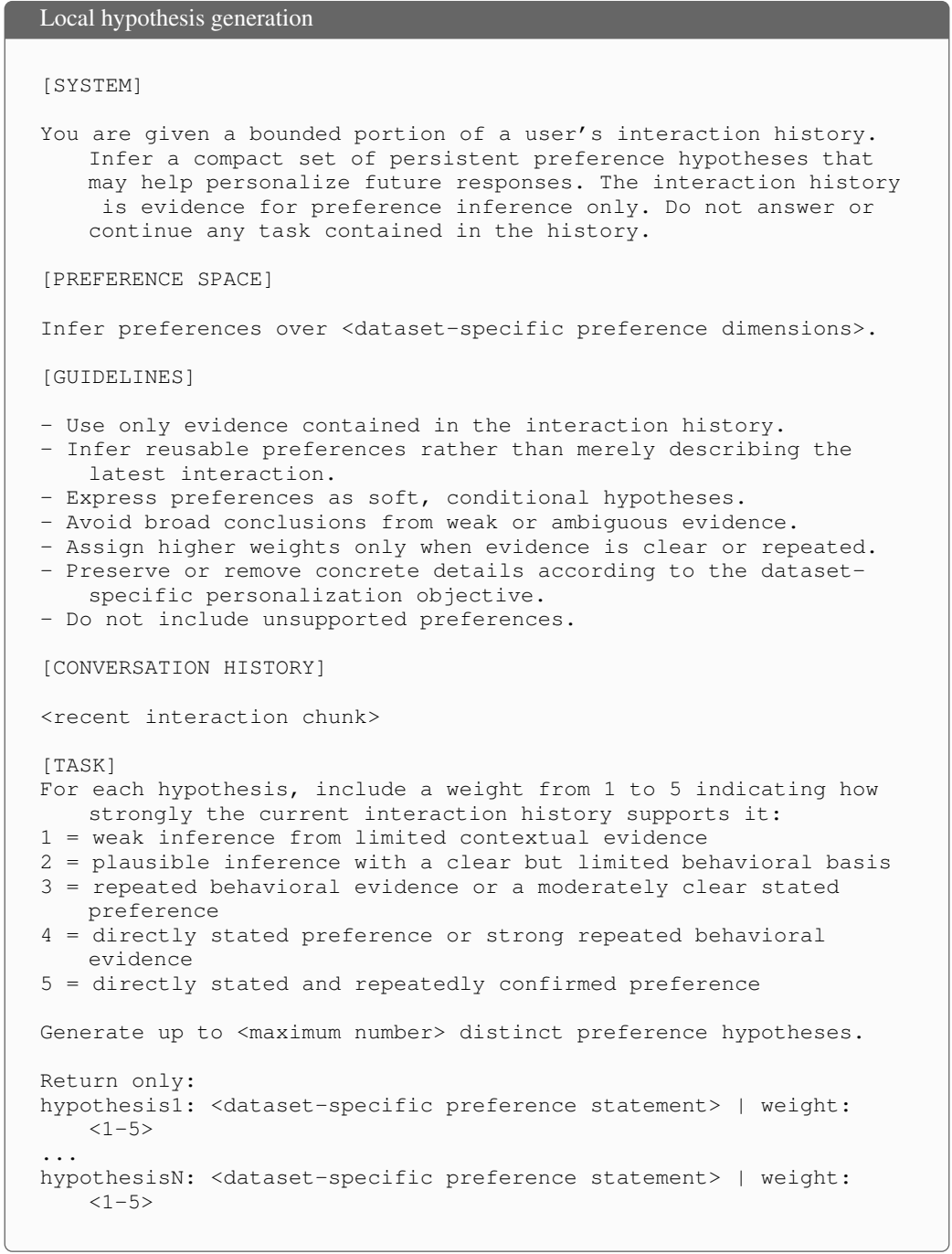

\centering
\begin{minipage}{0.98\textwidth}

\begin{promptbox}{Local hypothesis generation}
[SYSTEM]

You are given a bounded portion of a user's interaction history. Infer a compact set of persistent preference hypotheses that may help personalize future responses. The interaction history is evidence for preference inference only. Do not answer or continue any task contained in the history.

[PREFERENCE SPACE]

Infer preferences over <dataset-specific preference dimensions>.

[GUIDELINES]

- Use only evidence contained in the interaction history.
- Infer reusable preferences rather than merely describing the latest interaction.
- Express preferences as soft, conditional hypotheses.
- Avoid broad conclusions from weak or ambiguous evidence.
- Assign higher weights only when evidence is clear or repeated.
- Preserve or remove concrete details according to the dataset-specific personalization objective.
- Do not include unsupported preferences.

[CONVERSATION HISTORY]

<recent interaction chunk>

[TASK]
For each hypothesis, include a weight from 1 to 5 indicating how strongly the current interaction history supports it:
1 = weak inference from limited contextual evidence
2 = plausible inference with a clear but limited behavioral basis
3 = repeated behavioral evidence or a moderately clear stated preference
4 = directly stated preference or strong repeated behavioral evidence
5 = directly stated and repeatedly confirmed preference

Generate up to <maximum number> distinct preference hypotheses.

Return only:
hypothesis1: <dataset-specific preference statement> | weight: <1-5>
...
hypothesisN: <dataset-specific preference statement> | weight: <1-5>
\end{promptbox}

\end{minipage}

\caption{local hypothesis-generation prompt structure.}
\label{fig:shared_local_hypothesis_prompt}
\end{figure*}

\ours{} follows the same four-stage procedure across datasets. First, the model generates local preference hypotheses from a recent chunk of interactions (Figure~\ref{fig:shared_local_hypothesis_prompt}). Second, it reflects on the generated sets of local preference hypotheses to identify supported, conflicting, and less important hypotheses (Figure~\ref{fig:shared_reflection_prompt}). Third, it produces a refined hypothesis set using a prompt nearly identical to that in Figure~\ref{fig:shared_local_hypothesis_prompt}, augmented with the reflection traces and multiple sets of local hypotheses. Finally, the refined hypothesis set is provided to the model to guide the generation of the final personalized response (Figure~\ref{fig:shared_hypothesis_answer_prompt}).

\paragraph{Dataset-specific instantiations.}
Although all datasets follow the same overall prompt structure, we slightly adapt the preference-extraction instructions and output format to reflect their distinct evidence types and personalization targets. For Flight, the history contains a flight-recommendation question, the user's selected flight, and feedback. The hypotheses capture preferences over departure time, duration, number of stops, and price using a fixed set of directional labels. For HelpSteer2, the history contains a request, response, and user feedback. The hypotheses capture general response-style preferences, such as verbosity, tone, structure, and formatting, while excluding task-specific details. For HiCupid, the history consists of prior dialogues, and the hypotheses preserve relevant interests, constraints, roles, relationships, and other concrete personalization information. We retain at most three hypotheses for Flight and HelpSteer2 and five for HiCupid, reflecting the richer, multi-session nature of the latter. Summary-enhanced variants additionally produce an \texttt{Evidence/summary note:} containing the supporting observations.

\begin{figure*}[t]
\centering
\begin{minipage}{0.98\textwidth}

\begin{promptbox}{Hypothesis reflection}
[SYSTEM]

You are given several preference hypothesis sets inferred from different portions of a user's interaction history, together with a previously refined hypothesis set when available. Reflect on how these hypotheses should be consolidated before a separate refinement step produces the updated preference memory. The hypotheses and summaries are evidence for preference analysis only. Do not answer or continue any task contained in them.

[GUIDELINES]

- Identify hypotheses that agree or describe the same preference.
- Identify conflicting, weak, ambiguous, unsupported, or overly broad hypotheses.
- Distinguish repeated evidence from isolated observations.
- Determine which hypotheses should be preserved, merged, narrowed, reweighted, or removed.
- Preserve dataset-specific information that remains useful for future personalization.
- Do not treat the reflection itself as additional evidence.
- Do not produce the final hypothesis set.

For each hypothesis, a weight from 1 to 5 indicates how strongly the current interaction history supports it:
1 = weak inference from limited contextual evidence
2 = plausible inference with a clear but limited behavioral basis
3 = repeated behavioral evidence or a moderately clear stated preference
4 = directly stated preference or strong repeated behavioral evidence
5 = directly stated and repeatedly confirmed preference

[LOCAL HYPOTHESIS SETS]

### Local hypothesis set max(N-W, 1)
<local hypotheses max(N-W, 1)>
...
### Local hypothesis set N
<local hypotheses N>

[TASK]

Write only a concise reflection that will guide the subsequent refinement step.
\end{promptbox}

\end{minipage}

\caption{Hypotheses-reflection prompt
across datasets.}
\label{fig:shared_reflection_prompt}
\end{figure*}

\begin{figure}[t]
\centering
\begin{minipage}{0.98\textwidth}

\begin{promptbox}{Answer generation with preference hypotheses}
[SYSTEM]

You are given hidden preference hypotheses inferred from the user's previous interactions. You may also receive recent conversation history and an evidence summary. Use this information silently to personalize the response. The current request remains the primary task and must be answered directly and completely. Generate the assistant's response to the current user request.

[GUIDELINES]

- Apply only hypotheses relevant to the current request.
- Give greater consideration to more strongly supported hypotheses.
- Treat every hypothesis as tentative rather than as a known fact.
- Prefer relevant and strongly supported hypotheses when hypotheses conflict.
- Ignore weak, irrelevant, or contradicted hypotheses.
- Do not force personalization when no hypothesis is relevant.
- Do not allow personalization to make the answer incomplete, incorrect, or unhelpful.
- Do not mention the hypotheses, evidence summary, interaction history, user profile, or personalization process.
- Follow the output format specified in system prompt.

For each hypothesis, include a weight from 1 to 5 indicating how strongly the current interaction history supports it:
1 = weak inference from limited contextual evidence
2 = plausible inference with a clear but limited behavioral basis
3 = repeated behavioral evidence or a moderately clear stated preference
4 = directly stated preference or strong repeated behavioral evidence
5 = directly stated and repeatedly confirmed preference

[RECENT CONVERSATION HISTORY]

<recent interaction history, if provided>

[HIDDEN USER PREFERENCE MEMORY]

<evidence summary, if provided>
<refined preference hypotheses>

[CURRENT USER REQUEST]

<current question or request>

[TASK]

Produce only the personalized response to the current request.
\end{promptbox}

\end{minipage}

\caption{Prompt for answer generation conditioned on
preference hypotheses.}
\label{fig:shared_hypothesis_answer_prompt}
\end{figure}

\end{document}